\documentclass[review,11pt,times,authoryear]{elsarticle}
\usepackage[margin=1in]{geometry}
\usepackage{amssymb}
\usepackage{amsmath}
\usepackage{lineno}
\usepackage{graphicx}
\usepackage[colorlinks=true, allcolors=blue, breaklinks=true]{hyperref}
\usepackage{booktabs}
\usepackage{caption}
\usepackage{gensymb}
\usepackage{ragged2e}

\begin{document}

\begin{frontmatter}

    \title{\textbf{Investigating Robotaxi Crash Severity with Geographical Random Forest and the Urban Environment}}

    \author[a]{Junfeng Jiao}
    \author[a]{Seung Gyu Baik\corref{corr}}
    \author[a]{Seung Jun Choi\fnref{nau}}
    \author[a]{Yiming Xu}

    \affiliation[a]{{Urban Information Lab, The University of Texas at Austin}}
    \cortext[corr]{Corresponding Author (Seung Gyu Baik): Urban Information Lab, The University of Texas at Austin\\310 Inner Campus Drive, Austin, TX 78712, United States (\url{sgbaik@utexas.edu})}
    \fntext[nau]{Present Affiliation (Seung Jun Choi): Department of Geography, Planning, and Recreation, Northern Arizona University\\19 W McConnell Drive, Flagstaff, AZ 86011, United States}

    \begin{abstract}\indent
    \textbf{Introduction:} This paper quantitatively investigates the crash severity of Autonomous Vehicles (AVs) with spatially localized machine learning and macroscopic measures of the urban built environment. Extending beyond the microscopic effects of individual infrastructure elements, we focus on the city-scale land use and behavioral patterns, while addressing spatial heterogeneity and spatial autocorrelation. \textbf{Method:} We implemented a spatially localized machine learning technique called Geographical Random Forest (GRF) on the California AV collision dataset. Analyzing multiple urban measures, including points of interest, building footprint, and land use, we built a GRF model and visualized it as a crash severity risk map of San Francisco. \textbf{Results:} This paper presents three findings. First, spatially localized machine learning outperformed regular machine learning in predicting AV crash severity. The bias-variance tradeoff was evident as we adjusted the localization weight hyperparameter. Second, land use was the most important predictor, compared to intersections, building footprints, public transit stops, and Points Of Interest (POIs). Third, AV crashes were more likely to result in low-severity incidents in city center areas with greater diversity and commercial activities, than in residential neighborhoods. Residential land use is likely associated with higher severity due to human behavior and less restrictive environments. \textbf{Conclusion:} Predicting AV crash severity must incorporate spatial localization and land use. Counterintuitively, residential areas were associated with higher crash severity, compared to more complex areas such as commercial and mixed-use areas. \textbf{Practical Applications:} When robotaxi operators train their AV systems, it is recommended to consider the following. First, they should explicitly consider where their fleet operates and make localized algorithms for their perception system. Second, they should design safety measures specific to residential neighborhoods, such as slower driving speeds and more alert sensors.
    
    \end{abstract}
    
\end{frontmatter}

\newpage


\section{Introduction}

Autonomous Vehicles (AVs) and autonomous taxis, commonly known as \textit{robotaxi}, are already on the road. AVs have a fundamentally different motion planning paradigm compared to regular vehicles \citep{SAE-STD, NHTSA}, and their technology is anticipated to ``revolutionize the way consumers experience mobility \citep{Deichmann2023}''. AVs plan their movement by sensing the environment with various equipment like LiDAR, RADAR, cameras, ultrasonic, and GPS \citep{UM-CSS, Ignatious2022, Pan2024}. Integrating information from these sensors removes human factors from the loop and has a solid potential to improve transportation safety \citep{Bonnefon2016, RAND1}.

However, safety concerns remain regarding its commercial operations \citep{Fagnant2015, RAND2, Busch2024}. Also, there were a non-negligible number of fatal crash incidents that an AV could not avoid. Data from the National Highway Traffic Safety Administration (NHTSA) shows that there were 3566 vehicle crashes with automated driving systems (Level 3 or higher in Table \ref{table:sae-std}) during August 2021 to February 2025. Among them, 318 (9\%) crashes resulted in injuries, and 112 (3\%) were fatal \citep{NHTSA-DATA}. As it is projected that well over half of new car sales in the U.S., Europe, and China can be AVs by 2040 \citep{GoldmanSachs2024}, the number of injuries and fatalities can accordingly increase. In order to mitigate this, understanding the how and why of AV crashes is essential.

\begin{table}[h!]
	\begin{center}
            {\normalsize
		\begin{tabular}{clll} \toprule
                \textbf{Level} & \textbf{Description} & \multicolumn{2}{l}{\textbf{Responsibility}}\\ \cmidrule(lr){3-4}
                & & Vehicle & Human \\ \midrule
                0 & No Driving Automation & Momentary assistance & Drive\\
                1 & Driver Assistance & Assistance (brakes \textit{or} gas) & Drive\\
                2 & Partial Driving Automation & Assistance (brakes \textit{and} gas) & Drive\\
                3 & Conditional Driving Automation & Drive (if conditions are met) & Standby for takeover\\
                4 & High Driving Automation & Drive (in designated areas) & Ride as passenger\\
                5 & Full Driving Automation & Drive & Ride as passenger\\ \bottomrule
		\end{tabular}
            }
        \captionof{table}{SAE J3016 Standard for Driving Automation}
        \label{table:sae-std}
	\end{center}
\end{table}

A substantial amount of research on identifying factors associated with AV crashes has been conducted. Based on multiple exploratory data analyses, it is shown that crash occurrence is greatly related to pre-crash scenarios such as whether the vehicle was at an intersection or whether it was making a turn \citep{Favar2017, Xu2019, LiY2025}. Certain physical factors like lighting conditions and road conditions have also been identified to be associated with crash outcome \citep{Kurse2025, FuH2025}. This is understood to be caused by the microscopic interaction between AV sensors and the physical environment \citep{Vargas2021, Abdel-Aty2024}. However, few works have been attempted to examine the macroscopic pattern between AV crashes and the higher-level urban systems, like the built environment and land use.

The Built Environment (BE) is a comprehensive description of artificial surroundings built to support human activities: living, recreating, and working \citep{EPA-BE}. Examples include buildings, parks, sidewalks, commercial signage, shopfronts \citep{Portella2014}, open space, and utilities \citep{Hepp2016}. There are multiple BE measures relevant to transportation safety: intersections \citep{Flahaut2004} or roadway characteristics \citep{Wu2021}, street network connectivity \citep{Marshall2011}, population density \citep{Ferenchak2024}, development density \citep{Ewing2009}, and even the subjective perception of streetscape \citep{LiuY2025}. Land Use (LU), an overarching concept that illustrates economic and cultural activities practiced within the BE \citep{EPA-LU}, is another major aspect. LU has been utilized in previous non-AV studies either as area partitions \citep{Dumbaugh2010, Ukkusuri2012} or as specific locations \citep{LiuH2024, Kuo2024a}. Also, vehicle travel patterns are governed by how people use places \citep{Cervero1996, Yang2018, Choi2022}. This behavioral aspect cannot be captured by individual physical elements.

Broadening the scope from individual elements to macroscopic patterns of the BE and LU can be particularly useful in AV safety research because AVs are mobile, and their data is geographically distributed across multiple neighborhoods. The performance of AV perception systems, thus the data generating mechanism of AV crashes, can differ by location. This prohibits independent and identical distribution of data, and a model obtained from one location may not be able to perform well in another location \citep{Gao2023, Mai2025}. A macroscopic perspective opens a way to address this problem by incorporating \textit{spatial heterogeneity} and \textit{spatial autocorrelation}. Spatial heterogeneity (or \textit{spatial non-stationarity}) refers to the variation in parametric relationships across geographic space, while spatial autocorrelation (or \textit{spatial dependence}) describes the tendency of nearby locations to inherently exhibit similar characteristics \citep{Anselin1988, LeSage2009}. Some recent works have successfully incorporated spatially localized methods into their analyses on regular vehicles \citep{Wang2024, LiT2025}, but these efforts are yet to be extended to AVs and robotaxis.

This paper contributes to the literature in AV crash analysis from a macroscopic urban systems standpoint. We aim to reveal which BE measures and their corresponding LU behavior are associated with AV safety, and propose a novel spatially localized machine learning framework to predict crash outcomes. The remainder of the paper is structured as follows. First, we present a literature review focusing on how previous AV studies leave gaps regarding macroscopic BE patterns and spatial localization. Then, we organize the data, variables, and methodologies used in this study to predict safety outcomes (crash severity) and visualize their risk.

\section{Related Literature}

Existing works have extensively investigated the relationship between transportation safety and urban BE patterns. They attempted to identify BE/LU factors significantly associated with crash outcomes of non-AVs. \citet{Osama2017} examined BE, LU, socioeconomic factors, and road infrastructure on cyclist-motorist crashes. This paper found that crash occurrence increases with more commercial LU, and decreases with more recreational and residential LU. Similarly, \citet{Huang2018} found that commercial LU, road mileage, and intersection density are related to high occurrence crashes. Utilizing a Geographically Weighted Regression (GWR) model, this paper confirms spatial autocorrelation and heterogeneity in the BE-safety relationship. GWR had shown better predictive power than standard regression. A review by \citet{Merlin2020} emphasized the need for considering spatially localized data generating mechanisms. Their paper stated that it is unknown whether (1) specific designs are substantively safer or (2) different locations have different relationships. \citet{XiaoD2024} particularly addresses the Modifiable Areal Unit Problem (MAUP) \citep{Wong2009}, showing that geographical space partitioning based on traffic density is useful in identifying relevant BE features. Employment density, road density, highway density, LU diversity, and public transit accessibility were associated with higher crash occurrence. \citet{LiuH2024} used Google Street View images, Points Of Interest (POI) density, and a multi-level machine learning framework. The results indicated that educational POIs and corporate POIs were the greatest contributors to crash frequency. \citet{Ding2024} continue to investigate BE/LU factors and found that commercial LU, green areas, and bus stations were associated with greater crash occurrence. \citet{LiT2025}, via a geographically weighted neural network regressor, showed that intersection density and bus stop density have a greater effect on crash frequency than population density, LU diversity, and destination accessibility. The authors suggested that microscopic traffic interactions are more important than macroscopic BE factors.

These works on general transportation show that BE and LU are considered important factors in transportation safety. Examining both microscopic interactions and macroscopic patterns revealed that high-density urban development and commercial activities are less safe for vehicles. The need for such comprehensive analysis remains, especially for crash severity. \citet{Sarkar2025} stated in his paper that ``there is a lack of research studies investigating the role of ... traffic, non-traffic, physical, built-environment parameters'', regarding crash severity. Regarding research methods, spatial heterogeneity and spatial autocorrelation have been explicitly considered for modeling crash risks. Existing literature suggests the need for a spatially localized model for geographically distributed data like vehicle crashes.

However, existing literature focusing exclusively on AVs is often indifferent to these critical topics. The majority of works have weak consideration of the BE and its geospatial patterns. \citet{Favar2017} used AV collision reports from the California Department of Motor Vehicles (CA DMV) to visually reconstruct crashes and analyze their maneuvers. They showed that the crash occurrence of AVs correlates linearly with cumulative miles traveled. \citet{Xu2019} revisits the same data and adds satellite imagery to analyze road characteristics. Their model indicated that autonomous driving mode, intersections, curbside parking, rear-end collision, and one-way road types were significantly associated with crash severity. \citet{Kutela2022} extracted keywords with Natural Language Processing (NLP) and built a machine learning classifier to investigate AV crashes involving Vulnerable Road Users (VRUs). They concluded that crosswalks, intersections, traffic signals, and pre-crash maneuvers (turning, slowing down, stopping) were good predictors of VRU-related AV crashes. \citet{LiP2024} similarly extracted latent topics with NLP and machine learning. When investigating crash severity, controlling the class imbalance problem was highlighted to ensure classification (recall) power. \citet{Zhu2022} found that manufacturer, facility type, intersection, pre-crash movement, collision direction, lighting condition, and accident year are associated with crash severity. \citet{Kurse2025} and \citet{FuH2025} continue to update the literature with the CA DMV dataset and conclude that lighting, weather, and road surface conditions are major factors in AV crashes, emphasizing the need for reliable sensors. \citet{LiY2025} also updates the literature by utilizing CA DMV data from April 2018 to January 2024. They implemented a scenario-based program to analyze AV crashes. \citet{ChenH2020} analyzed POI data from 4 categories: commercial, residential, office, and traffic. With this, they quantified development intensity around the crash sites (buffer analysis) by calculating POI diversity. The authors concluded that dense developments and LU diversity cause complex environments, in which AVs can malfunction. \citet{Ren2022} included more specified variables: metro stops, trees, LU, parks, and schools. They found that rain, mixed-use LU, the presence of light rail, and driving at night were associated with higher AV crash severity. \citet{Kuo2024b} gathered a much wider range of POIs, such as community centers, nightclubs, hospitals, ice cream shops, and parking spaces. Their machine learning model showed that entertainment facilities, places of worship, fountains, and nightclubs were positively associated with AV crash severity.

The literature review above presents three takeaways. First, the majority of AV crash analyses naively focused on the microscopic effects of individual elements. They did not sufficiently consider the macroscopic, behavioral patterns of the urban systems. Research on AV safety should also investigate city-scale BE patterns beyond roadway design and pre-crash maneuvers. Second, existing AV crash studies did not explicitly incorporate spatial heterogeneity or autocorrelation. To close the gap, there is a need to bring spatially localized methodologies from non-AV studies and apply them to AV studies.

\section{Materials and Methods}

\subsection{Data}

\begin{figure}[h!]
    \centering
    \includegraphics[width=1\linewidth]{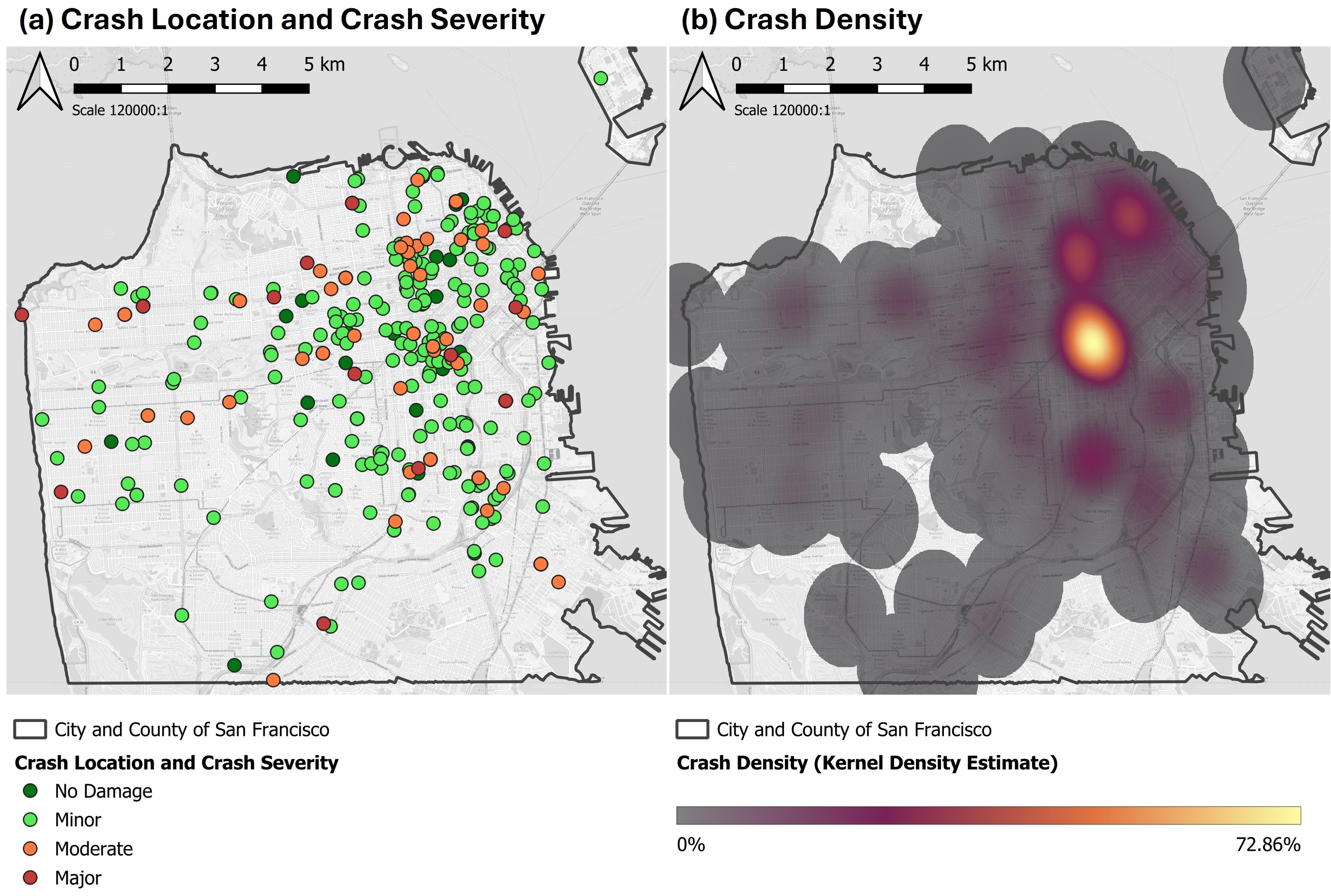}
    \caption{(a) Crash Location and Crash Severity; (b) Crash Density}
    \label{fig:studyarea}
\end{figure}

We used the CA DMV dataset, which had been utilized in many previous AV safety studies. We utilized a Large Language Model (LLM) to extract information from California's Report of Traffic Accident Involving an Autonomous Vehicle (OL 316) documents. We limited our time frame from 2019 to 2024, and our study area to the City and County of San Francisco. The location, severity, and kernel density estimate (KDE) of 496 AV crashes are shown in Figure \ref{fig:studyarea}. The data was clustered around the downtown area.

\begin{figure}[p!]
    \centering
    \includegraphics[width=0.9\linewidth]{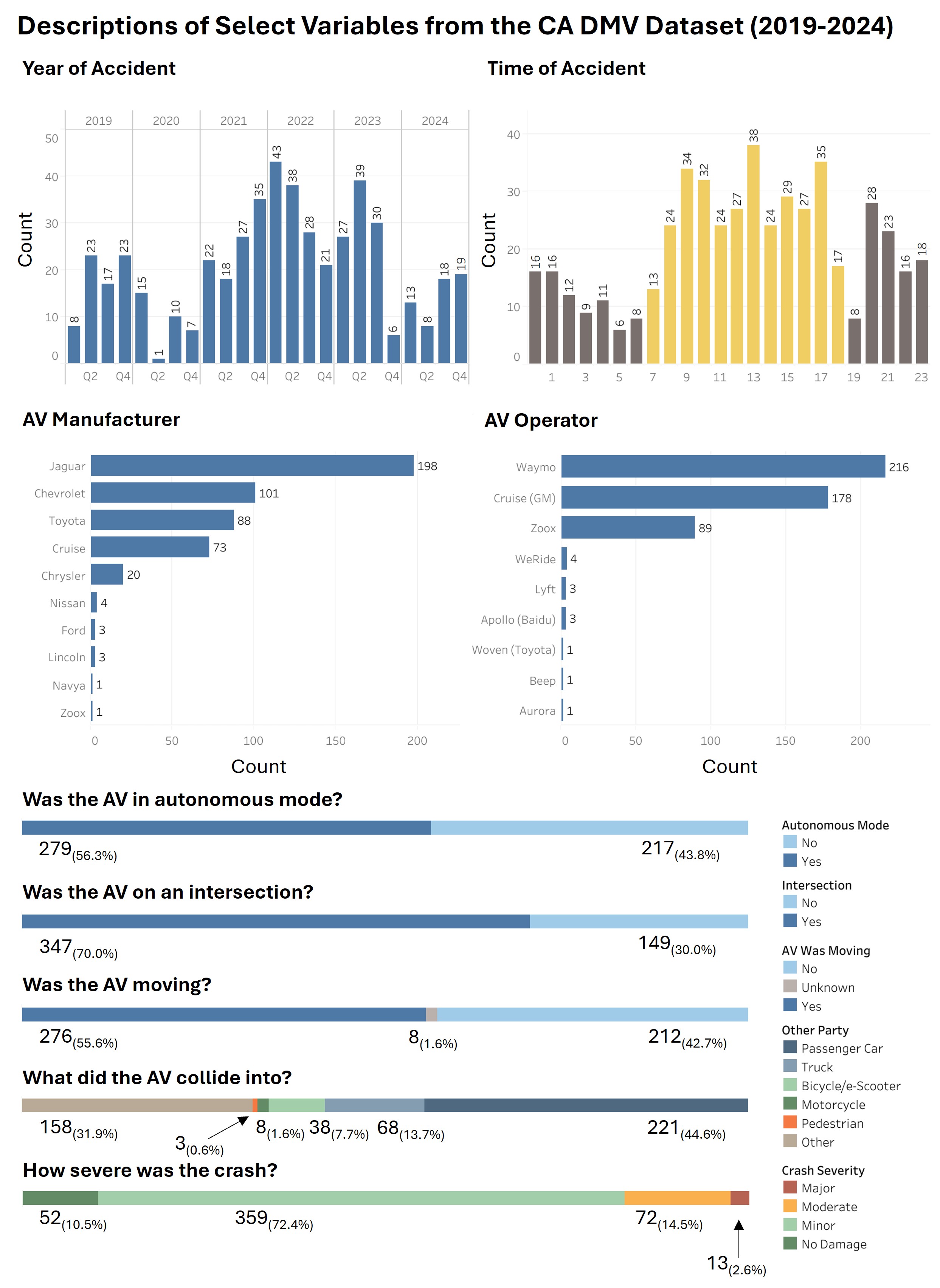}
    \caption{Description of Select Variables from the CA DMV Dataset (2019-2024)}
    \label{fig:collage}
\end{figure}

Descriptions of a few select variables are in Figure \ref{fig:collage}. Accidents were most frequent between 2021 and 2023, with a noticeable drop in 2020 Q2 and 2023 Q4. Accidents happened throughout the day except at sunrise and sunset. The biggest AV manufacturer was Jaguar, followed by Chevrolet and Toyota. The biggest business operating the AVs was Waymo, followed by Cruise and Zoox. Among 496 accidents, 43.8\% happened while autonomous driving was disengaged. 70.0\% of accidents happened at an intersection, and 55.6\% happened while the AV was moving. The AVs most frequently collided with another passenger car at 44.6\%. The crash severity was greatly imbalanced as nearly 83\% were \textit{Minor} or \textit{No Damage} instances.

\subsection{Variables}

\begin{figure}[h!]
    \centering
    \includegraphics[width=1\linewidth]{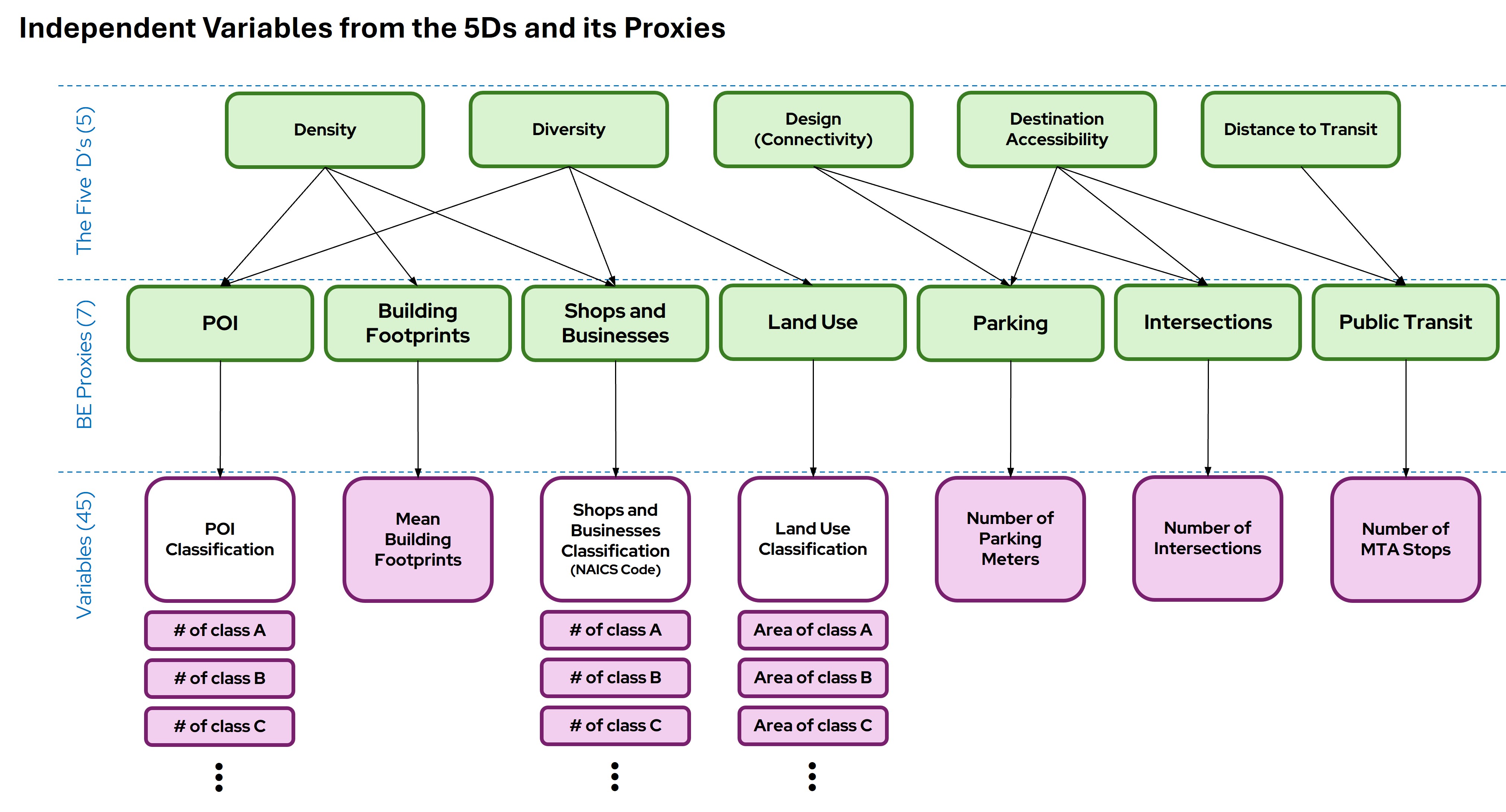}
    \caption{Independent Variables from the 5Ds and its Proxies}
    \label{fig:variables}
\end{figure}

We select BE variables that can capture the macroscopic, city-scale BE patterns and use them to predict AV crash outcomes. While there is no clear definition of what is microscopic or macroscopic, we implement the `5D' framework. The 5D framework is an extension of the `3D' or `4D' framework \citep{Cervero1997, Ewing2001, Ewing2009}, and is a well-established way to represent high-level BE/LU characteristics in transportation studies. In order to quantitatively model the BE-safety relationship, we set BE proxies that comprehensively transcribe the 5Ds while generating tabular data, as Figure \ref{fig:variables}. There are two reasons for setting proxies. First, the five `D's themselves are not individual variables and can be interpreted in multiple interwoven ways \citep{Handy2018}. Second, no such index or metric directly representing the 5Ds is available in public data sources.

\renewcommand{\arraystretch}{0.66}

\begin{table}[p!]
\begin{center}
{\footnotesize
\begin{tabular}{llllrrrr} \toprule
\textbf{BE Proxy} & \textbf{Variable} & \textbf{Description} & \textbf{Unit} & \multicolumn{4}{c}{\textbf{Summary Statistics}}\\ \cmidrule(lr){5-8}
&&&& Min & Max & Mean & SD\\ \midrule
      Points Of & \texttt{POI Sust} & Sustenance (Food \& Beverage) POIs & \textit{Count} & 0 & 393 & 95.55 & 72.37\\
Interests (POI) & \texttt{POI Edu} & Education POIs & \textit{Count} & 0 & 9 & 2.46 & 3.94 \\
                & \texttt{POI Trans} & Transportation POIs & \textit{Count} & 0 & 25 & 4.89 & 3.81\\
                & \texttt{POI Fin} & Financial POIs & \textit{Count} & 0 & 146 & 10.37 & 17.46\\
                & \texttt{POI Health} & Healthcare POIs & \textit{Count} & 0 & 16 & 2.89 & 2.59\\
                & \texttt{POI EAC} & Entertainment, Arts, \& Culture POIs & \textit{Count} & 0 & 17 & 3.32 & 3.94\\
                & \texttt{POI Pub} & Public Service POIs & \textit{Count} & 0 & 25 & 4.89 & 3.81\\
                & \texttt{POI Faci} & Facilities POIs & \textit{Count} & 0 & 146 & 10.37 & 17.46\\
                & \texttt{POI Waste} & Waste Management POIs & \textit{Count} & 0 & 63 & 4.16 & 6.52\\
                & \texttt{POI Other} & Other POIs & \textit{Count} & 0 & 53 & 6.55 & 8.61\\ \cmidrule{1-8}
Bldg. Footprint & \texttt{Mean BFP} & Mean Building FootPrint & $\text{ft}^2$ & 94 & 2078 & 404.63 & 344.17\\ \cmidrule{1-8}
      Shops and & \texttt{Shop Acc} & Accommodation & \textit{Count} & 0 & 144 & 41.80 & 25.34\\
     Businesses & \texttt{Shop Admin} & Administrative \& Support Service & \textit{Count} & 0 & 393 & 25.08 & 37.67\\
                & \texttt{Shop AER} & Arts, Entertainment, \& Recreation & \textit{Count} & 2 & 162 & 51.59 & 29.66\\
                & \texttt{Shop Cert} & Other certain service & \textit{Count} & 0 & 169 & 29.27 & 24.14\\
                & \texttt{Shop Const} & Construction & \textit{Count} & 0 & 134 & 35.76 & 17.64\\
                & \texttt{Shop Fin} & Financial Service & \textit{Count} & 0 & 1149 & 36.72 & 111.66\\
                & \texttt{Shop Food} & Food Service & \textit{Count} & 0 & 463 & 111.07 & 86.28\\
                & \texttt{Shop Info} & Information Service & \textit{Count} & 0 & 603 & 34.86 & 61.61\\
                & \texttt{Shop Insu} & Insurance Service & \textit{Count} & 0 & 180 & 4.43 & 14.26\\
                & \texttt{Shop Manu} & Manufacturing & \textit{Count} & 0 & 96 & 11.50 & 11.68\\
                & \texttt{Shop PEH} & Private Education \& Health Serv. & \textit{Count} & 0 & 430 & 58.17 & 68.63\\
                & \texttt{Shop PST} & Professional, Scientific, \& Tech. Serv. & \textit{Count} & 0 & 3143 & 168.20 & 283.87\\
                & \texttt{Shop RERL} & Real Estate, Rental \& Leasing Serv. & \textit{Count} & 0 & 1945 & 183.37 & 173.00\\
                & \texttt{Shop Retail} & Retail Trade & \textit{Count} & 0 & 535 & 91.95 & 76.29\\
                & \texttt{Shop Trans} & Transportation \& Warehousing & \textit{Count} & 0 & 53 & 13.62 & 10.78\\
                & \texttt{Shop Util} & Utilities & \textit{Count} & 0 & 16 & 1.38 & 2.32\\
                & \texttt{Shop Whole} & Wholesale Trade & \textit{Count} & 0 & 110 & 15.98 & 15.38\\
                & \texttt{Shop Multi} & Multiple Classification & \textit{Count} & 0 & 196 & 33.52 & 24.22\\
                & \texttt{Shop Other} & Other unspecified businesses & \textit{Count} & 3 & 6618 & 824.21 & 690.40\\ \cmidrule{1-8}
       Land Use & \texttt{LU Resident} & Residential & $\text{km}^2$ & 0 &0.3290 &0.1115 &0.0859\\
                & \texttt{LU MixRes} & Mixed Use with Residential & $\text{km}^2$ & 0 &0.2618 &0.0685 &0.0397\\
                & \texttt{LU Mixed} & Mixed Use without Residential & $\text{km}^2$ & 0 &0.6072 &0.0410 &0.0512\\
                & \texttt{LU CIE} & Cultural, Institutional, \& Education & $\text{km}^2$ & 0 &0.3830 &0.0194 &0.0274\\
                & \texttt{LU PDR} & Production, Distribution, \& Repair & $\text{km}^2$ & 0 &0.0105 &0.0080 &0.0119\\
                & \texttt{LU Medi} & Medical & $\text{km}^2$ & 0 &0.0487 &0.0029 &0.0067\\
                & \texttt{LU Visit} & Hotels \& Motels & $\text{km}^2$ & 0 &0.0488 &0.0027 &0.0069\\
                & \texttt{LU MIPS} & Mgmt., Info., \& Professional Serv. & $\text{km}^2$ & 0 &0.3219 &0.0349 &0.0442\\
                & \texttt{LU RetailEnt} & Retail \& Entertainment & $\text{km}^2$ & 0 &0.1098 &0.0206 &0.0228\\
                & \texttt{LU Openspace} & Open Space & $\text{km}^2$ & 0 & 10.9373 &0.3290 & 1.4143\\
                & \texttt{LU Vacant} & Vacant or undeveloped lots & $\text{km}^2$ & 0 & 2.8961 &0.0648 &0.1412\\
                & \texttt{LU Other} & Other unspecified or unknown LU & $\text{km}^2$ & 0 &0.2398 &0.0038 &0.0243\\ \cmidrule{1-8}
        Parking & \texttt{Parking Meters} & SFMTA Parking Meters & \textit{Count} & 0 & 1222 & 396.23 & 340.32\\ \cmidrule{1-8}
   Intersection & \texttt{Intersections} & Intersection of two or more roads & \textit{Count} & 2 & 91 & 34.80 & 16.18\\ \cmidrule{1-8}
 Public Transit & \texttt{MTA Stops} & SFMTA Public Transit Stops & \textit{Count} & 0 & 35 & 13.33 & 7.90\\ \bottomrule
\end{tabular}
}
\captionof{table}{Shortlisted BE Variables}
\label{table:variablestable}
\end{center}
\end{table}

The shortlisted 45 independent variables from 7 proxies are in Table \ref{table:variablestable}. Descriptions and methodologies for each variable are as follows. (1) POI data has been previously used for building machine learning models that predict crash severity. The categorization of POIs can capture diversity \citep{ChenH2020} while the types of individual POIs can be an independent variable itself \citep{Kuo2024b}. The number of POIs in proximity to crash sites can indicate density. In this paper, we gathered OpenStreetMap POI data via Overpass Turbo. We used the query \texttt{amenity=*} to extract all available POI features. In total, there were 10 POI categories. All other data are retrieved from the San Francisco Open Data Portal, DataSF. (2) Building footprint is the area of land each building occupies. Building footprint is a good indicator of density, which has been investigated in classical studies \citep{Ewing2009, Dumbaugh2009}. It can also differentiate between small-scale and large-scale installations \citep{Huang2018}. The dataset we used is individual building geometry built from airborne LiDAR scanning. We calculated the surface area and its centroid of each building via GIS. (3) Shops and businesses are individual locations of registered, tax-paying businesses. The dataset we accessed was last updated on April 13, 2025. Shops and businesses indicate the density and diversity of the crash site, along with POIs. Most businesses were tagged with North American Industry Classification System (NAICS) codes, which enables the classification of businesses. (4) Land use is the current status and nature of how a parcel of land is being used. This is different from zoning regulations that define how land can be used within designated zoning districts. Land use has been used in previous studies \citep{Osama2017, Wu2021} and supports the diversity element. The DataSF land use dataset (2023) had 12 land use classifications. (5) For parking and public transit, we used the data organized by the San Francisco Municipal Transportation Agency (SFMTA) system, which was available on DataSF. (6) The locations of intersections were obtained from road network data. We extracted all points where two or more roads intersect via GIS.

\subsection{Quantifying AV Crash Severity Risk}

\begin{figure}[h!]
    \centering
    \includegraphics[width=0.9\linewidth]{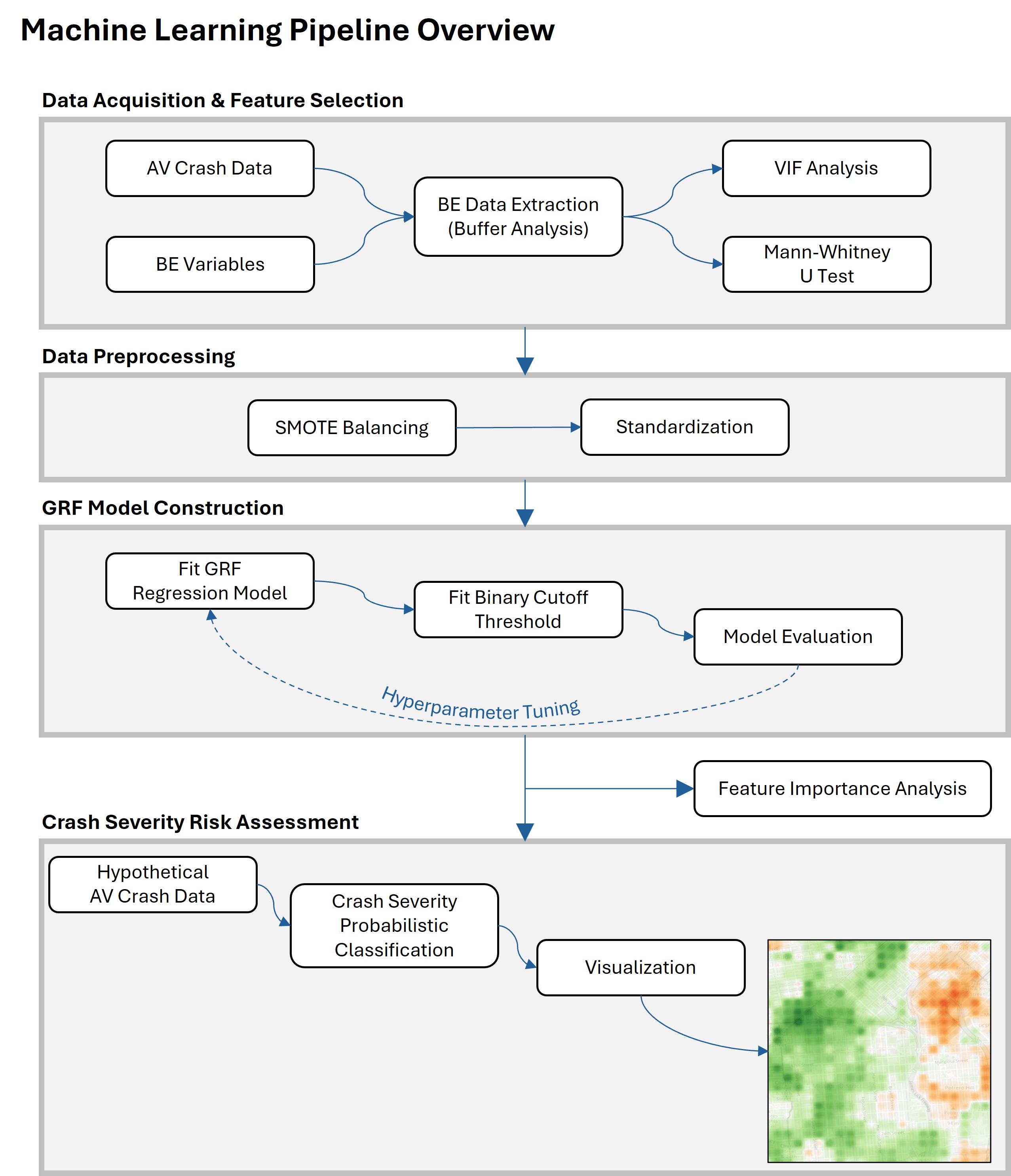}
    \caption{Machine Learning Pipeline Overview}
    \label{fig:overview}
\end{figure}

We implemented a machine learning model to quantify AV crash severity risk at a given location. An overview of the pipeline is as Figure \ref{fig:overview}. As the first step, we conducted a buffer analysis to acquire tabular data for each shortlisted BE variables. Then, we performed feature selection for our machine learning model with a set a rules.

For our buffer analysis, we chose the size of the buffer as follows. According to \citet{Ignatious2022}, who collected specifications of multiple AV sensors, the max perception range of AV LiDARs ranged from 30 meters to 245 meters. Since we are interested in the macroscopic BE/LU beyond the AVs' physical interaction, the size of our buffer had to be larger than this. According to \citet{LiuX2023}, who tried different buffer sizes for machine learning, a buffer of 600 meters showed the best results in predicting metro ridership with BE. Since we are interested in road vehicles and their data points are much granular, our buffer size had to be smaller than this. A common buffer size in urban planning is \textit{Perry's Neighborhood Unit} of $1/4$ miles \citep{Perry1929, Mehaffy2015}. This value is roughly 400 meters, which is between 245 and 600 meters. It has also been used in many previous transportation studies \citep{Cervero2011, PSRC2015, Ewing2016}. As a result, we have chosen our buffer size as 400 meters. We collected values for each independent variable in Table \ref{table:variablestable} within 400 meters of each crash location.

Feature selection was performed via Mann-Whitney $U$ tests (MWU) and Variance Inflation Factor (VIF) analysis to reduce dimensionality while retaining useful features. MWU is a robust, non-parametric (rank-ordered) counterpart to the $t$-test and is used to determine whether two independent samples come from the same distribution or not \citep{Montgomery2018}. A $t$-test was inappropriate since it requires approximately Gaussian distributions, homogeneous variance, and similar sample sizes between the (crash severity) categories \citep{Kim2019}. The resulting $p$-values from MWU are the probability of the two samples coming from two indistinguishable distributions. In our study, we reassigned the crash severity into \textit{High} and \textit{Low} to perform a pairwise MWU. VIF detects multicollinearity of each variable. It measures how much it inflates a linear model's regression coefficients by fitting one variable with all other variables. High VIF can make the model redundantly complex and less interpretable \citep{James2021}. Previous works have removed variables with VIF $>10$ \citep{Ren2022, Kuo2024b}. Deleting a high VIF variable is acceptable because a high VIF means that other collinear variables can work as a proxy of the deleted variable. In our study, we used variables that are: (1) VIF $<10$ \textit{or} (2) MWU $p<0.2$.

Next, we addressed the class imbalance problem, which is a known issue in the CA DMV dataset. Our AV crash data has a great imbalance among the 4 severity levels as shown in Figure \ref{fig:variables}. Class imbalance leads to the lack of representative samples in the high severity class, making it difficult to build a general and accurate predictive model \citep{HeH2009, Krawczyk2016}. To mitigate this, we used the Synthetic Minority Oversampling Technique (SMOTE) to artificially create data points from the minority class \citep{Chawla2002}. SMOTE has been used in previous geospatial research \citep{Johnson2016, Zhang2019} and AV crash analysis \citep{Kuo2024b}. For each feature vector (one crash instance) $\mathbf{x}$ in the minority class sample, SMOTE creates an artificial sample $\mathbf{x_{NEW}}$ along the line joining $\mathbf{x}$ and one of its nearest neighbors $\mathbf{x_{N}}$ as Equation \ref{smote}. The weight parameter $w$ is random between 0 and 1. In addition, since we have diverse BE variables, their numeric scales are also diverse. For example, the scale of the mean building footprint (square feet) is about one hundred times bigger than the scale of the number of intersections (count). For this reason, we standardized all variables into their standard score (z-score).

\begin{gather}
\mathbf{x_{NEW}} = \mathbf{x} + w(\mathbf{x_{N}}-\mathbf{x})
\label{smote}
\end{gather}

Our spatially localized machine learning model, the Geographical Random Forest (GRF), is a unique configuration of Random Forest (RF). RF is a popular ensemble learning technique with injected randomness regarding which features to use \citep{Breiman2001}. RF is effective in prediction tasks with geospatial data \citep{Du2015, Amani2020}. From the training data $D$, RF randomly draws $B$ samples with replacement and makes multiple bootstrap subsets ($D_1, D_2, D_3, \dots, D_b, \dots D_B$). An intermediate predictor (a decision tree) is trained independently for each bootstrap subset. For each feature vector (one crash instance) $\mathbf{x}$, a hypothesis (crash severity prediction) $\hat{y}_{\text{RF}}$ is made as Equations \ref{rf1} and \ref{rf2}.

\begin{gather} 
\mathbf{x} \xrightarrow{\quad \text{Prediction with } D_{b} \quad} \hat{y}_{b} \qquad \forall b \in {1,2,3,\dots} 
\label{rf1} \\
\frac{1}{B} \sum_{b=1}^B \hat{y}_{b} = \hat{y}_{\text{RF}}
\label{rf2}
\end{gather}

GRF builds multiple locally calibrated RF models, one `forest' per location. These RF sub-models are trained using a limited number of nearby observations. This can directly address spatial heterogeneity and autocorrelation \citep{Georganos2021}. GRF is also known as Geographically Weighted Random Forest (GWRF) and has been utilized in environmental science \citep{Aguirre-Gutierrez2021, Sailaja2024, LuW2025} and transportation studies \citep{Wu2024}. For each feature vector (one crash instance) $\mathbf{x}$ at location $i$, a local hypothesis (crash severity prediction) $\hat{y}_{\text{RF}(i)}$ is made as Equation \ref{grf2}. $(u_i,v_i)$ is the geographic coordinate of location $i$. Localized predictions are combined with global (regular) RF as Equation \ref{grf3}. $\hat{y}_{\text{GRF}(i)}$ is the final prediction for crash instance $\mathbf{x}$ in location $i$. GRF has two special hyperparameters: the bandwidth parameter $n$ and the localization weight parameter $a$. Parameter $n$ is the number of nearest neighbors used to define each neighborhood (kernel). Parameter $a$ balances between local RF and global RF. This controls how `spatially aware' or `localized' the GRF model is. If $a=1$, GRF predictions are completely localized (lower spatial bias). If $a=0$, GRF predictions are the same as a regular, global RF (lower spatial variance).

\begin{gather}
\mathbf{x} \xrightarrow{\quad \text{Global RF} \quad} \hat{y}_{\text{RF}}
\label{grf1} \\
\mathbf{x}, (u_i,v_i) \xrightarrow{\quad \text{Local RF at location } i \text{ with } n \text{ neighbors} \quad} \hat{y}_{\text{RF}(i)}
\label{grf2} \\
\hat{y}_{\text{GRF}(i)} = (a)\hat{y}_{\text{RF}(i)} + (1-a)\hat{y}_{\text{RF}}
\label{grf3}
\end{gather}

In order to quantify the crash severity risk of a location, we formulated a binary probabilistic classification problem. Assuming that a crash has happened at an arbitrary location, we attempted to predict its probability of being a high-severity crash. We reassigned 1 (high-severity) for \textit{Moderate} and \textit{Major} instances and 0 (low-severity) for \textit{No Damage} and \textit{Minor} instances. This made our GRF regression model give out predictions between 0 and 1. We used a recently developed Python library called PyGRF \citep{Sun2024} to implement GRF. For evaluation purposes, we temporarily discretized GRF's predictions to either 0 or 1, and calculated performance metrics as Equations \ref{recall} to \ref{acc}.

{\small
\begin{gather}
\text{Recall} = \frac{\text{True Positives(TP)}}{\text{True Positives(TP)}+\text{False Negatives(FN)}}
\label{recall}\\
\text{Precision} = \frac{\text{True Positives(TP)}}{\text{True Positives(TP)}+\text{False Positives(FP)}}
\label{prec}\\
\text{Accuracy} = \frac{\text{TP}+\text{TN}}{\text{TP}+\text{FN}+\text{TN}+\text{FP}}
\label{acc}
\end{gather}
}

We chose Accuracy (Equation \ref{acc}) as our GRF model's primary performance metric. Accuracy enables one to evaluate the model's overall predictive power with no preference for either high severity or low severity. We performed hyperparameter tuning to obtain: (1) a best accuracy model, (2) a best recall model, (3) a completely global RF model, and (4) a completely local RF model. For this, we created 4,913 hypothetical crash locations across the study area, 100 meters apart in all cardinal directions. A buffer analysis was conducted to extract the same variables as the original CA DMV dataset. This synthetic crash dataset was fed into the trained best accuracy GRF model to quantify the probability of high-severity crashes across San Francisco. Resulting predictions were visualized via Inverse Distance Weighing (IDW).

In addition to interpreting the GRF model and evaluating each BE feature's association with AV crash severity, we conducted a feature importance analysis. For each variable and each node split in the GRF model, we calculated the average reduction of Gini impurity. In our study, Gini impurity measures each node's diversity (`impureness') of corresponding crash severity values. For each branch split, an important feature will reduce this impurity by effectively differentiating high-severity and low-severity crashes. A less important feature will less reduce this impurity. Averaging out the reduction of Gini impurity across all nodes in a tree, we get the Gini importance of individual features. In order to reveal the direction of association, we conducted a series of \textit{t}-tests between high-severity zones and low-severity zones.

\section{Results}

\subsection{Feature Selection Results}

Feature selection results are in Table \ref{table:featselection}. MWU tests revealed no statistically significant differences between the two (\textit{High} vs \textit{Low}) severity classes, except \texttt{POI Other}. A few variables showed a trivially weak association ($p<0.2$). VIF tests found strong multicollinearity (VIF $>10$) in over half of the Shops and Businesses variables. This was expected, as urbanization often leads to the clustering of businesses; where one kind of shop is found, others are likely to follow. Applying our feature selection criteria of (1) VIF $<10$ \textit{or} (2) MWU $p<0.2$, 24 out of 45 variables were used to train our GRF model.

\begin{table}[h!]
\begin{center}

{\footnotesize
\begin{tabular}{lllrrrc} \toprule
\textbf{BE Proxy}& \textbf{Variable} & \multicolumn{3}{c}{\textbf{Mann-Whitney U}} & \textbf{VIF} & \textbf{Selected}\\ \cmidrule{3-5}
& & Values greater in: & U & \textit{p} & & \\ \midrule
            POI & \texttt{POI Sust} &&17527.5&0.41&52.08&\\
                & \texttt{POI Edu} &&17999.5&0.27&\textsuperscript{\textdagger}5.67&$\checkmark$\\
                & \texttt{POI Trans} &&17358.5&0.47&11.07&\\
                & \texttt{POI Fin} &&17288&0.49&18.62&\\
                & \texttt{POI Health} &&17403.5&0.45&\textsuperscript{\textdagger}7.46&$\checkmark$\\
                & \texttt{POI EAC} &&18092&0.24&10.03&$\checkmark$\\
                & \texttt{POI Pub} &&17723&0.35&16.32&\\
                & \texttt{POI Faci} &&17288.5&0.43&\textsuperscript{\textdagger}6.49&$\checkmark$\\
                & \texttt{POI Waste} &&17728&0.35&\textsuperscript{\textdagger}6.96&$\checkmark$\\
                & \texttt{POI Other} &High Severity&19352.5&\textsuperscript{*}0.04& \textsuperscript{\textdagger}8.46&$\checkmark$\\\cmidrule{1-7}
Bldg. Footprint & \texttt{Mean BFP} &&17493&0.42&\textsuperscript{\textdagger}7.12&$\checkmark$\\\cmidrule{1-7}
      Shops and & \texttt{Shop Acc} &&17533&0.41&17.21&\\
     Businesses & \texttt{Shop Admin} &&17367&0.47&31.80&\\
                & \texttt{Shop AER} &&17552.5&0.40&30.23&\\
                & \texttt{Shop Cert} &&17648&0.37&38.56&\\
                & \texttt{Shop Const} &&18016.5&0.26&27.22&\\
                & \texttt{Shop Fin} &&17723&0.35&78.86&\\
                & \texttt{Shop Food} &&18148.5&0.23&40.78&\\
                & \texttt{Shop Info} &&17919&0.29&47.91&\\
                & \texttt{Shop Insu} &&17780.5&0.33&32.92&\\
                & \texttt{Shop Manu} &&17665.5&0.37&\textsuperscript{\textdagger}8.58&$\checkmark$\\
                & \texttt{Shop PEH} &&18168&0.22&10.36&\\
                & \texttt{Shop PST} &&17786.5&0.33&376.38\\
                & \texttt{Shop RERL} &High Severity&18582.0&$^\circ$0.13&10.31&$\checkmark$\\
                & \texttt{Shop Retail} &&17351&0.47&30.96&\\
                & \texttt{Shop Trans} &&17917.5&0.29&12.46&\\
                & \texttt{Shop Util} &Low Severity&19014.0&$^\circ$0.07&\textsuperscript{\textdagger}4.63&$\checkmark$\\
                & \texttt{Shop Whole} &High Severity&18334.5&$^\circ$0.19&15.54&$\checkmark$\\
                & \texttt{Shop Multi} &&17407&0.45&54.54&\\
                & \texttt{Shop Other} &&17633.5&0.38&349.26&\\\cmidrule{1-7}
       Land Use & \texttt{LU Resident} &&17733.5&0.35&\textsuperscript{\textdagger}8.11&$\checkmark$\\
                & \texttt{LU MixRes} &High Severity&18688.5&$^\circ$0.12&12.57&$\checkmark$\\
                & \texttt{LU Mixed} &Low Severity&18393.0&$^\circ$0.17&\textsuperscript{\textdagger}2.97&$\checkmark$\\
                & \texttt{LU CIE} &&17604&0.39&\textsuperscript{\textdagger}1.78&$\checkmark$\\
                & \texttt{LU PDR} &&17750.5&0.34&\textsuperscript{\textdagger}4.69&$\checkmark$\\
                & \texttt{LU Medi} &&17947&0.28&\textsuperscript{\textdagger}1.70&$\checkmark$\\
                & \texttt{LU Visit} &&18286&0.20&\textsuperscript{\textdagger}4.96&$\checkmark$\\
                & \texttt{LU MIPS} &&17448&0.44&12.22&\\
                & \texttt{LU RetailEnt} &&17905&0.30&\textsuperscript{\textdagger}7.16&$\checkmark$\\
                & \texttt{LU Openspace} &&18087&0.25&\textsuperscript{\textdagger}1.41&$\checkmark$\\
                & \texttt{LU Vacant} &Low Severity&18548.5&$^\circ$0.14&\textsuperscript{\textdagger}2.40&$\checkmark$\\
                & \texttt{LU Other} &&17838.5&0.32&\textsuperscript{\textdagger}1.42&$\checkmark$\\\cmidrule{1-7}
        Parking & \texttt{Parking Meters} &&17540.5&0.41&26.16&\\\cmidrule{1-7}
   Intersection & \texttt{Intersections} &Low Severity&18758.0&$^\circ$0.11&13.56&$\checkmark$\\\cmidrule{1-7}
 Public Transit & \texttt{MTA Stops} &&17632&0.38&14.79&$\checkmark$\\ \bottomrule
                &&&&{\footnotesize$^\circ$:$p<0.2$}&{\footnotesize*:$p<0.05$}&{\footnotesize\textsuperscript{\textdagger}:VIF$<10$}
\end{tabular}
}
\captionof{table}{Feature Selection Results}
\label{table:featselection}
\end{center}
\end{table}

\subsection{Geographical Random Forest Model}

The performances of our GRF models are as Table \ref{table:performance}. The best accuracy ($a=0.50$) model had 78\% overall accuracy and 53\% recall. The best recall ($a=0.16$) model had 61\% accuracy and 82\% recall. Varying localization parameter $a$ showed no change in average $R^2$, which is how much a model can explain the variability in data. Extreme localization like $a=0.01$ or $a=0.99$ decreased overall accuracy. Completely localized RF showed better accuracy and lower recall when compared with completely global (regular) RF.

\begin{table}[h!]
\begin{center}

{\small
\begin{tabular}{cccccccc} \toprule
        \textbf{Parameter $a$} & \multicolumn{2}{c}{\textbf{Model Weight}} & \textbf{$R^2$} & \textbf{Accuracy} & \textbf{Precision} & \textbf{Recall} & \textbf{Remarks}\\ \cmidrule{2-3}
        & {\footnotesize Global RF (\%)} & {\footnotesize Local RF (\%)} & & & & & \\ \midrule
       0.01 & 99 & 1 &0.67&0.65&0.30&0.76&\\
       0.16 & 84 & 16 &0.67&0.61&0.28&\textbf{0.82}&Best Recall\\
       0.25 & 75 & 25 &0.67&0.73&0.33&0.59&\\
       0.50 & 50 & 50 &0.67&\textbf{0.78}&0.39&0.53&Best Accuracy\\
       0.75 & 25 & 75 &0.67&0.71&0.33&0.71&\\
       0.99 & 1 & 99 &0.67&0.72&0.34&0.71&\\\bottomrule
\end{tabular}
}
\captionof{table}{GRF Model Perfomance}
\label{table:performance}
\end{center}
\end{table}

\begin{figure}[h!]
    \centering
    \includegraphics[width=1\linewidth]{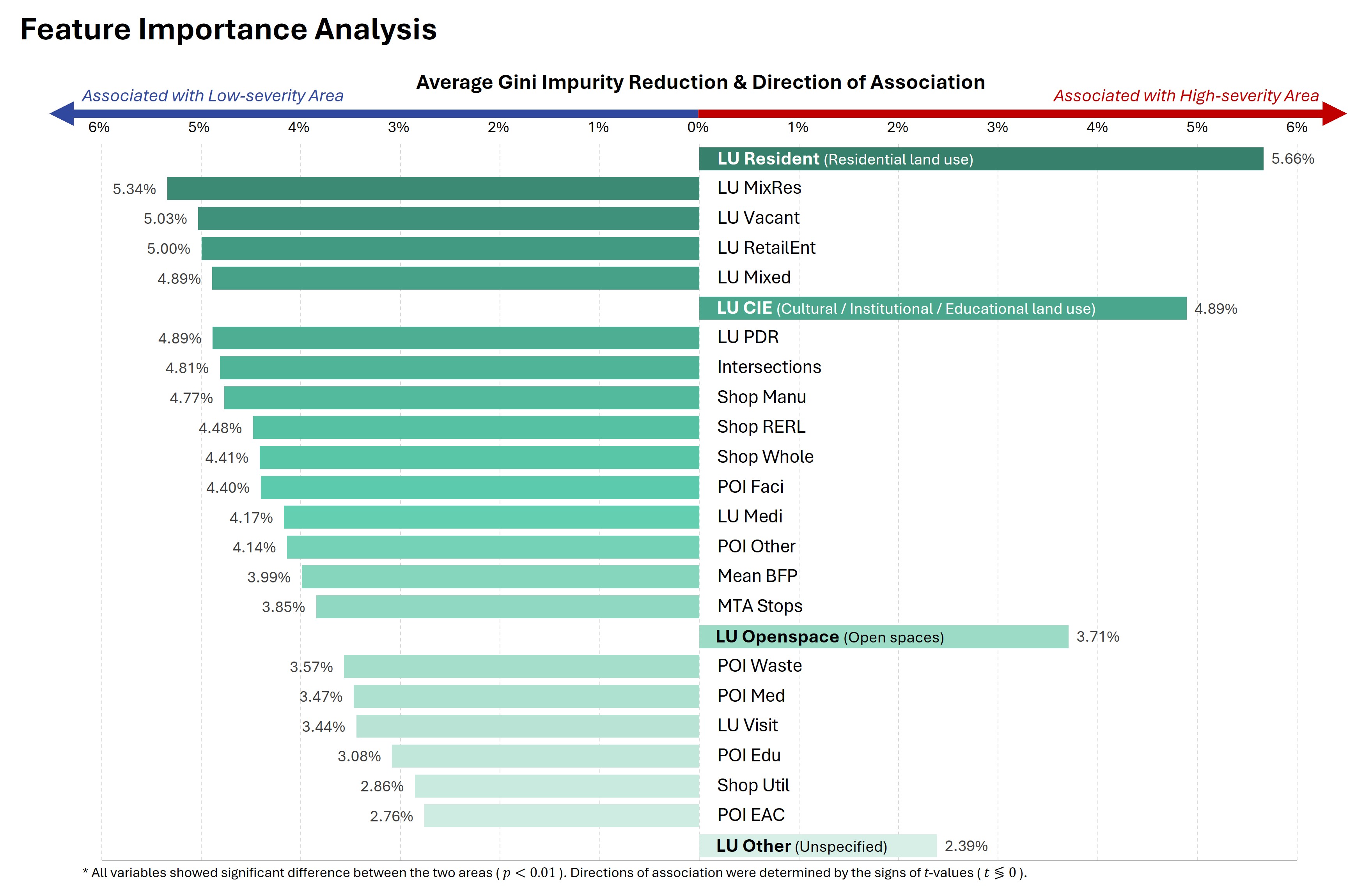}
    \caption{Feature Importance Analysis}
    \label{fig:gini}
\end{figure}

Feature importance analysis results are as Figure \ref{fig:gini}. Land use was deemed more important than intersections, shops, and POIs. Particularly, \texttt{LU Resident} and \texttt{LU MixRes} were ranked the highest, reducing Gini impurity by 5.66\% and 5.34\%, respectively. \texttt{LU Other} was least important at 2.39\%. POIs, building footprint, and Public Transit were relatively unimportant. The directions of association show that residential land use (\texttt{LU Resident}), cultural, institutional, or educational land use (\texttt{LU CIE}), and open spaces (\texttt{LU Openspace}) caused the GRF model to predict a higher AV crash severity risk. On the other hand, mixed or commercial land use (\texttt{LU MixRes}, \texttt{LU Mixed}, \& \texttt{LU RetailEnt}) was associated with lower risk.

\subsection{AV Crash Severity Risk Map}

\begin{figure}[h!]
    \centering
    \includegraphics[width=1\linewidth]{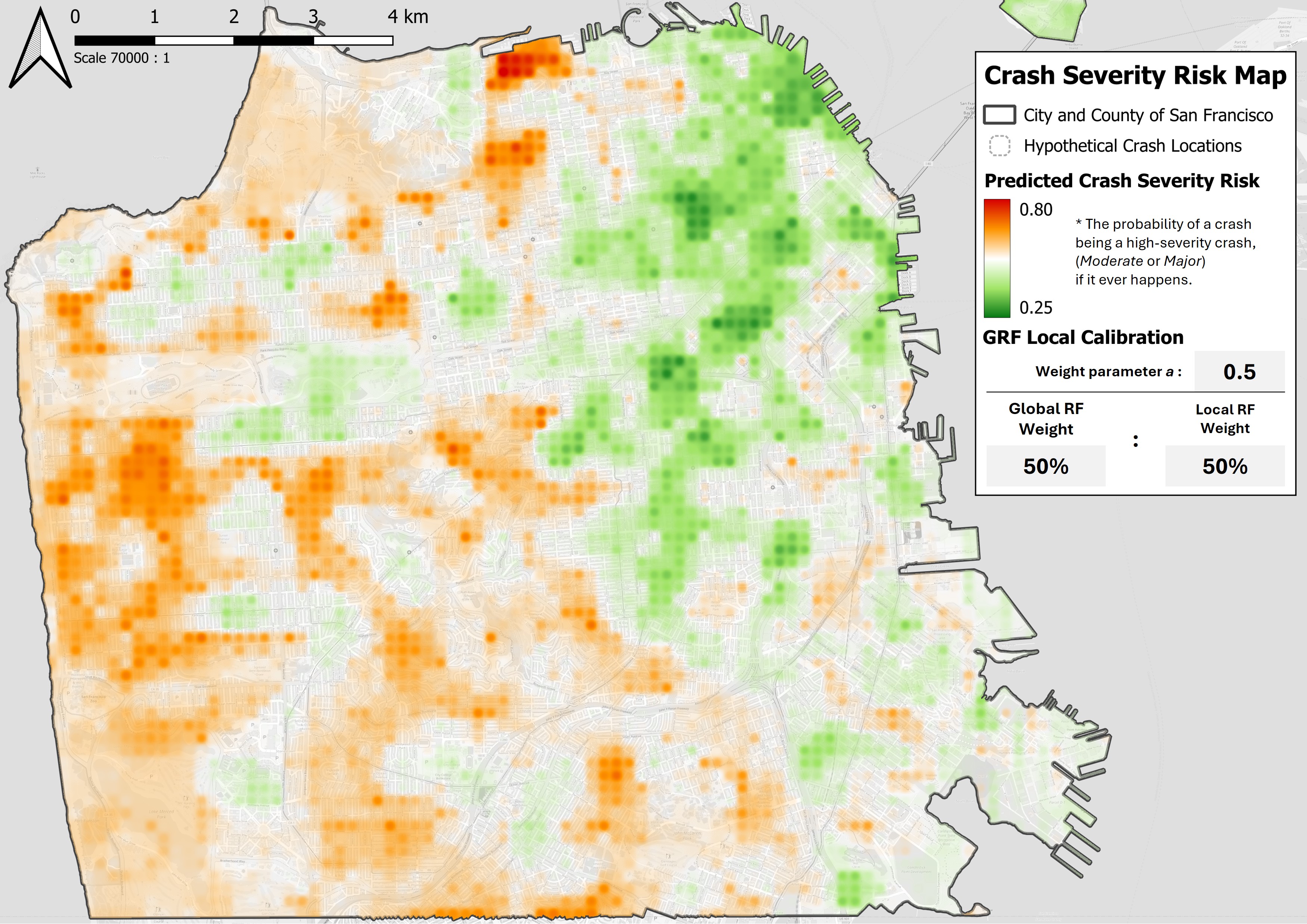}
    \caption{Crash Severity Risk Map of San Francisco}
    \label{fig:baselineviz}
\end{figure}

 We quantified and visualized the AV crash severity risk of San Francisco. The resulting map from our best accuracy ($a=0.5$) GRF model is as Figure \ref{fig:baselineviz}. Green cells are where high-severity crashes are predicted to be less likely, and orange cells are where high-severity crashes are more likely. In general, high-severity predictions were more frequent in western residential areas than in northeastern downtown areas. Distinct patches were identified throughout the study area.

\section{Discussion}

\subsection{Effects of Geographical Random Forest}

We could perform a controlled comparison between global RF and localized RF. When all else remains the same, adjusting the weight parameter $a$ significantly impacts model performance. Despite different localization weights similarly explaining the variability in data ($R^2$), it was possible for GRF models ($a>0.01$) to show better performance than regular RF ($a=0.01$) models, in either overall accuracy (50\% localization) or recall (16\% localization). This suggests that AV crash severity exhibits spatial heterogeneity and spatial autocorrelation. Geographical localization improves a machine learning model's predictive power by capturing these spatial effects. However, predicting AV crash severity with GRF worked best if we balance between the global model and local models. Appropriately controlling the level of localization was essential.

\begin{figure}[h!]
    \centering
    \includegraphics[width=1\linewidth]{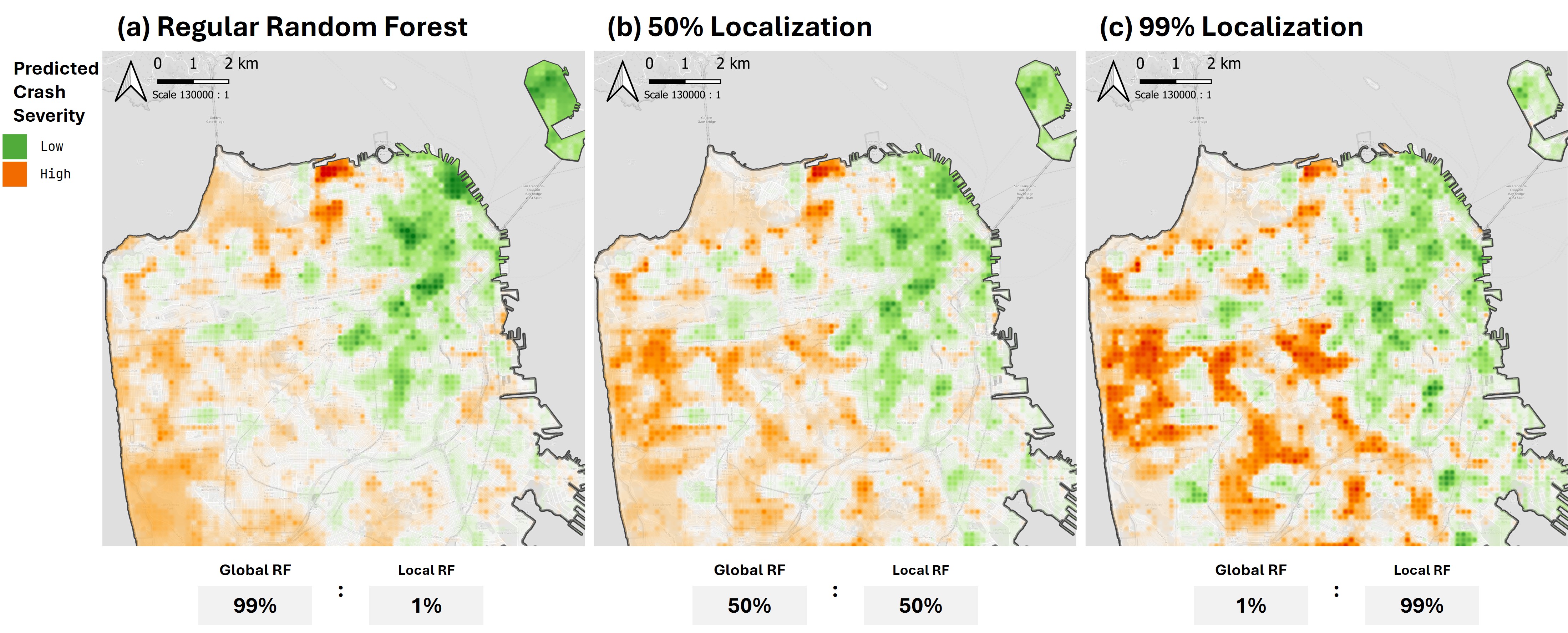}
    \caption{AV Crash Severity Risk Profiles with Varying Localization Weights; (a) Regular Random Forest; (b) 50\% Localization; (c): 99\% Localization}
    \label{fig:comparison}
\end{figure}

Changing the parameter $a$ yielded varying results as Figure \ref{fig:comparison}. Model (b) at 50\% localization is the best accuracy model. Model (a) is the visualization made with 1\% weight on local RF, and model (c) is made with 99\% weight on local RF. In general, stronger localization yielded higher and more pronounced predictions on AV crash severity. Model (c) showed more extreme predictions, especially towards higher severity. Also, predicted low-severity `safe zones' were more granular and scattered across the study area.

We can interpret this from a bias-variance tradeoff perspective. GRF is designed to suppress variance with a global model and suppress bias with local models \citep{Georganos2021}. Our AV crash severity risk profiles in Figure \ref{fig:comparison} show that stronger localization (c) leads to increased variance in crash severity risks, while weaker localization (a) leads to superficial and coarse predictions. Considering that Figure \ref{fig:comparison} (b) had the best overall accuracy at 78\%, we can say that Figure \ref{fig:comparison} (a) resulted from underfitting and Figure \ref{fig:comparison} (c) resulted from overfitting. Underfitting means that the model did not sufficiently capture locally heterogeneous signals, and overfitting means that the model failed to generalize by focusing too much on local signals.

\subsection{Land Use and AV Crash Severity}

The feature importance analysis in Figure \ref{fig:gini} shows that LU classification was the most important BE measure in predicting AV crash severity. The number of intersections, the number of shops, POIs, and the mean building footprints turned out to be relatively weak predictors. This suggests the need for AV systems to be aware of their higher-level surroundings. On top of recognizing individual physical elements, the perception systems of AVs need to be fine-tuned regarding which part of the city the system is expected to run on. Figure \ref{fig:gini} also shows that residential LU is associated with higher AV crash severity, while commercial or mixed-use LU is associated with the opposite.

\begin{figure}[h!]
    \centering
    \includegraphics[width=0.9\linewidth]{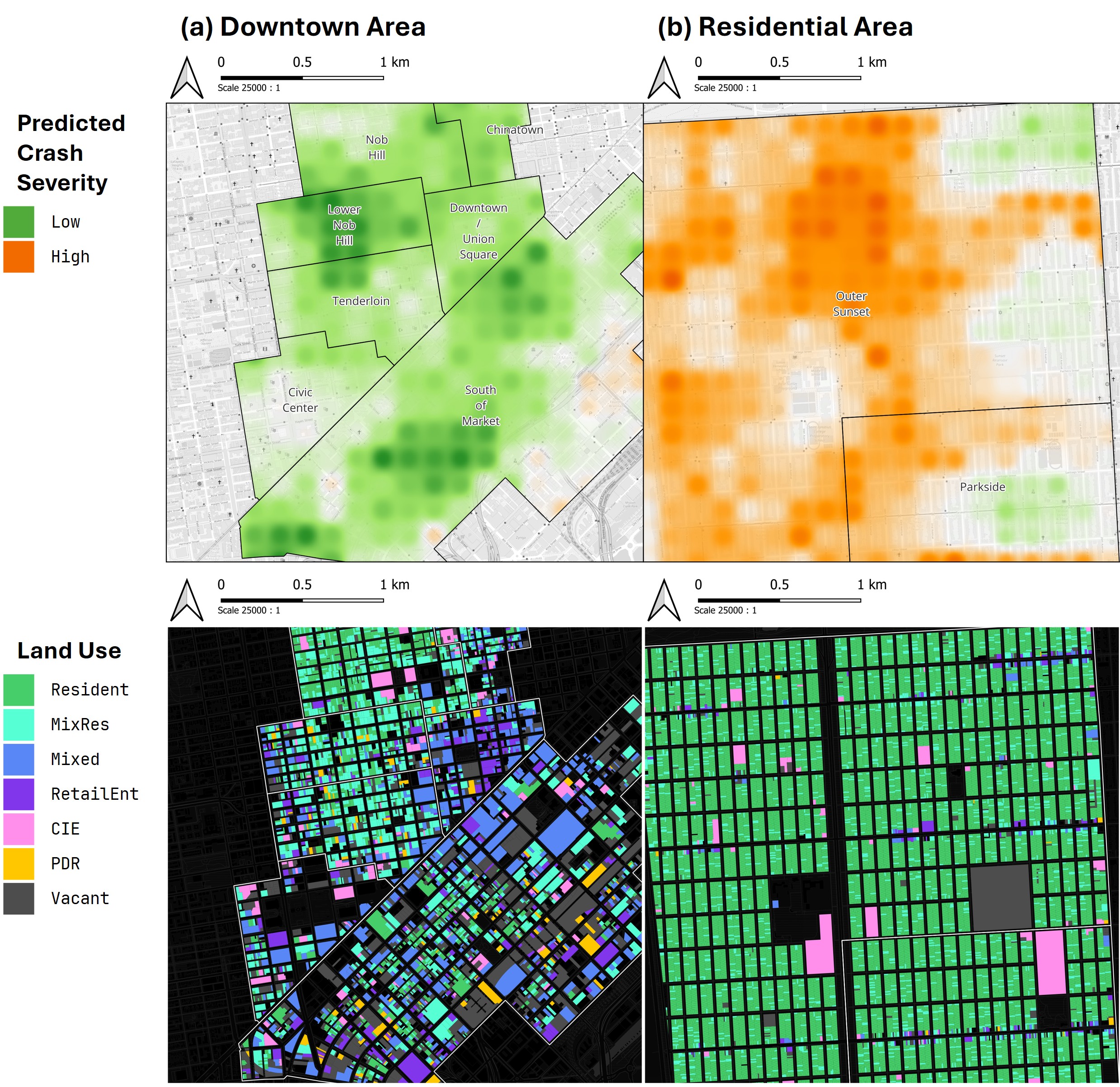}
    \caption{Land Use Comparison; (a) Downtown Area; (b) Residential Area}
    \label{fig:discussion}
\end{figure}

In order to further support this finding, we conducted a neighborhood-level examination as Figure \ref{fig:discussion}. The crash severity predictions in the downtown area (a) and a residential area (b) alongside parcel-level land use classification (seven highest feature importance only) are shown. LU diversity and commercial activity were noticeably greater (\texttt{MixRes}, \texttt{Mixed}, and \texttt{RetailEnt} LU) at low-severity locations like Lower Nob Hill and South of Market (SoMa). On the other hand, high-severity locations were dominated by uniform, rectangular tract homes (\texttt{Residential} LU). Also, the three small, mixed-use strip malls on the east side of Outer Sunset and Parkside showed distinct low-severity clusters surrounded by high-severity predictions. Despite vehicle crashes happening more frequently in diverse commercial areas \citep{Osama2017, Huang2018}, our analyses show that uniform residential areas can pose a greater risk to AV crash severity. Since crash frequency and crash severity are not the same, we can interpret this as follows. Once an AV crash happens in a residential area, it is expected to be more severe, compared to commercial areas.

There are two possible reasons for this. First, residential LU may be related to certain human behaviors that lead to high-severity crashes. For instance, pedestrians in residential areas jaywalk more often, and this can be a critical cause for AV crashes \citep{Zhang2025}. AVs are less likely to react promptly if pedestrians are not expected on the roads. Also, certain pedestrians, including children and joggers, are more likely to exhibit active behaviors and less likely to promptly react. This is supported by the positive association of education (\texttt{LU CIE}) and open spaces (\texttt{LU Openspace}) with higher crash severity. Second, residential LU may lead AVs to exhibit certain driving characteristics that can result in high-severity crashes. The complex, congested environment in dense city centers limits the reliability of AV perception systems \citep{ChenH2020}, naturally restricting their speed and freedom of movement. On the contrary, in residential areas, especially where development density is low, there is less congestion and fewer traffic interactions. This means that AVs can drive at higher speeds. Higher speed, in turn, leaves less time for the system to react to unforeseen obstacles. A less restrictive environment may be paradoxically causing more severe crash outcomes.

\section{Conclusions}

This study predicts the crash severity of Autonomous Vehicles (AVs) with Geographical Random Forest (GRF) and city-scale measures of the urban built environment. This study contributes to the growing body of research on AV safety from two novel aspects. First, our spatially localized machine learning model incorporates spatial heterogeneity and spatial autocorrelation in predicting AV crash severity. Second, we investigate the higher-level, city-scale built environment measures, including both Points Of Interest (POIs) and land use classification, in predicting AV crash severity. Neither aspect has been addressed in existing AV safety research.

This study has three major findings. First, our GRF model reveals that a spatially localized machine learning technique may perform better than regular machine learning techniques. It also highlights the importance of carefully controlling the degree of localization from a bias-variance tradeoff standpoint. Second, land use classification was most important in predicting AV crash severity, among other measures like intersections, building footprints, public transit stops, and POIs. Third, diverse and complex commercial areas were associated with lower AV crash severity, whereas residential neighborhoods were associated with higher AV crash severity. From these findings, it is suggested that residential land use affects AV crash severity via: (1) certain behavioral patterns of pedestrians and (2) a less restrictive, less complex environment resulting in higher AV speeds.

These findings offer two practical insights for commercial operations of AVs, especially autonomous taxi services (robotaxi). First, robotaxi operators should incorporate geographic location when training the perception systems for their AV fleet. The system will perform better when the model is localized; a set of motion planning rules that differ across geographic space can capture the latent effects from spatial heterogeneity and spatial autocorrelation. Second, even if dense city centers are more complex and diverse, it does not mean that residential neighborhoods are safer. More robust safety measures should be implemented when driving in residential neighborhoods, because if a crash ever happens, its outcome is expected to be more severe. This can include lower speed limits, more consideration of potential unsafe human behavior, and better training on such environments, such as fine-tuning the AVs' perception system to be more alert.

There are limitations that should be addressed in subsequent research. First, our study focused exclusively on San Francisco, which has unique urban characteristics that may not generalize to other cities. Despite San Francisco having the most accessible dataset, a cross-city analysis is required to test our findings. Second, we examined crash severity without controlling for crash frequency and traffic volume, which limits our understanding of overall AV safety management. Since crash severity does not directly relate to human injury, preventing a crash from happening in the first place must also be investigated. Interpreting low-severity locations as `safe streets' is not perfectly appropriate. Integrating both crash severity and crash occurrence should present a more comprehensive AV safety risk assessment.

\newpage

\section*{CRediT authorship contribution statement}
{\small \textbf{Junfeng Jiao:} Supervision, Project administration, Funding acquisition, Conceptualization. \textbf{Seung Gyu Baik:} Writing - Original Draft, Conceptualization, Methodology, Software, Visualization. \textbf{Seung Jun Choi:} Writing - Review \& Editing, Supervision, Conceptualization, Investigation. \textbf{Yiming Xu:} Writing - Review \& Editing, Resources, Data Curation, Validation.
}

\section*{Funding}
This work was supported by the \href{https://bridgingbarriers.utexas.edu/good-systems}{UT Good Systems Grand Challenge}; the National Science Foundation [grant numbers 2043060, 2133302, 1952193, 2125858, and 2236305]; the USDOT Consortium for Cooperative Mobility for Competitive Megaregions; and The MITRE Corporation.

\section*{Data availability}
Autonomous Vehicle Collision Reports used in this study are publicly available on the California Department of Motor Vehicles website({\footnotesize \url{www.dmv.ca.gov/portal/vehicle-industry-services/autonomous-vehicles}}).


\bibliographystyle{model5-names} 

\begin{FlushLeft}
{\small
\bibliography{main.bib}

\begin{thebibliography}{84}
\expandafter\ifx\csname natexlab\endcsname\relax\def\natexlab#1{#1}\fi
\providecommand{\url}[1]{\texttt{#1}}
\providecommand{\href}[2]{#2}
\providecommand{\path}[1]{#1}
\providecommand{\DOIprefix}{doi:}
\providecommand{\ArXivprefix}{arXiv:}
\providecommand{\URLprefix}{URL: }
\providecommand{\Pubmedprefix}{pmid:}
\providecommand{\doi}[1]{\href{http://dx.doi.org/#1}{\path{#1}}}
\providecommand{\Pubmed}[1]{\href{pmid:#1}{\path{#1}}}
\providecommand{\bibinfo}[2]{#2}
\ifx\xfnm\relax \def\xfnm[#1]{\unskip,\space#1}\fi
\bibitem[{Abdel-Aty \& Ding(2024)}]{Abdel-Aty2024}
\bibinfo{author}{Abdel-Aty, M.}, \& \bibinfo{author}{Ding, S.} (\bibinfo{year}{2024}).
\newblock \bibinfo{title}{A matched case-control analysis of autonomous vs human-driven vehicle accidents}.
\newblock {\it \bibinfo{journal}{Nat. Commun.}\/},  {\it \bibinfo{volume}{15}\/}, \bibinfo{pages}{4931}. \DOIprefix\doi{10.1038/s41467-024-48526-4}.
\bibitem[{Aguirre-Gutiérrez et~al.(2021)Aguirre-Gutiérrez, Rifai, Shenkin, Oliveras, Bentley, Svátek, Girardin, Both, Riutta, Berenguer, Kissling, Bauman, Raab, Moore, Farfan-Rios, Figueiredo, Reis, Ndong, Ondo, Bengone, Mihindou, de~Seixas, Adu-Bredu, Abernethy, Asner, Barlow, Burslem, Coomes, Cernusak, Dargie, Enquist, Ewers, Ferreira, Jeffery, Joly, Lewis, Marimon-Junior, Martin, Morandi, Phillips, Quesada, Salinas, Marimon, Silman, Teh, White \& Malhi}]{Aguirre-Gutierrez2021}
\bibinfo{author}{Aguirre-Gutiérrez, J.}, \bibinfo{author}{Rifai, S.}, \bibinfo{author}{Shenkin, A.}, \bibinfo{author}{Oliveras, I.}, \bibinfo{author}{Bentley, L.~P.}, \bibinfo{author}{Svátek, M.}, \bibinfo{author}{Girardin, C.~A.}, \bibinfo{author}{Both, S.}, \bibinfo{author}{Riutta, T.}, \bibinfo{author}{Berenguer, E.}, \bibinfo{author}{Kissling, W.~D.}, \bibinfo{author}{Bauman, D.}, \bibinfo{author}{Raab, N.}, \bibinfo{author}{Moore, S.}, \bibinfo{author}{Farfan-Rios, W.}, \bibinfo{author}{Figueiredo, A. E.~S.}, \bibinfo{author}{Reis, S.~M.}, \bibinfo{author}{Ndong, J.~E.}, \bibinfo{author}{Ondo, F.~E.}, \bibinfo{author}{Bengone, N.~N.}, \bibinfo{author}{Mihindou, V.}, \bibinfo{author}{de~Seixas, M. M.~M.}, \bibinfo{author}{Adu-Bredu, S.}, \bibinfo{author}{Abernethy, K.}, \bibinfo{author}{Asner, G.~P.}, \bibinfo{author}{Barlow, J.}, \bibinfo{author}{Burslem, D.~F.}, \bibinfo{author}{Coomes, D.~A.}, \bibinfo{author}{Cernusak, L.~A.}, \bibinfo{author}{Dargie, G.~C.}, \bibinfo{author}{Enquist, B.~J.},
  \bibinfo{author}{Ewers, R.~M.}, \bibinfo{author}{Ferreira, J.}, \bibinfo{author}{Jeffery, K.~J.}, \bibinfo{author}{Joly, C.~A.}, \bibinfo{author}{Lewis, S.~L.}, \bibinfo{author}{Marimon-Junior, B.~H.}, \bibinfo{author}{Martin, R.~E.}, \bibinfo{author}{Morandi, P.~S.}, \bibinfo{author}{Phillips, O.~L.}, \bibinfo{author}{Quesada, C.~A.}, \bibinfo{author}{Salinas, N.}, \bibinfo{author}{Marimon, B.~S.}, \bibinfo{author}{Silman, M.}, \bibinfo{author}{Teh, Y.~A.}, \bibinfo{author}{White, L.~J.}, \& \bibinfo{author}{Malhi, Y.} (\bibinfo{year}{2021}).
\newblock \bibinfo{title}{Pantropical modelling of canopy functional traits using sentinel-2 remote sensing data}.
\newblock {\it \bibinfo{journal}{Remote Sens. Environ.}\/},  {\it \bibinfo{volume}{252}\/}, \bibinfo{pages}{112122}. \DOIprefix\doi{10.1016/j.rse.2020.112122}.
\bibitem[{Amani et~al.(2020)Amani, Ghorbanian, Ahmadi, Kakooei, Moghimi, Mirmazloumi, Moghaddam, Mahdavi, Ghahremanloo, Parsian, Wu \& Brisco}]{Amani2020}
\bibinfo{author}{Amani, M.}, \bibinfo{author}{Ghorbanian, A.}, \bibinfo{author}{Ahmadi, S.~A.}, \bibinfo{author}{Kakooei, M.}, \bibinfo{author}{Moghimi, A.}, \bibinfo{author}{Mirmazloumi, S.~M.}, \bibinfo{author}{Moghaddam, S. H.~A.}, \bibinfo{author}{Mahdavi, S.}, \bibinfo{author}{Ghahremanloo, M.}, \bibinfo{author}{Parsian, S.}, \bibinfo{author}{Wu, Q.}, \& \bibinfo{author}{Brisco, B.} (\bibinfo{year}{2020}).
\newblock \bibinfo{title}{Google earth engine cloud computing platform for remote sensing big data applications: A comprehensive review}.
\newblock {\it \bibinfo{journal}{IEEE J. Sel. Top. Appl. Earth. Obs. Remote Sens.}\/},  {\it \bibinfo{volume}{13}\/}, \bibinfo{pages}{5326--5350}. \DOIprefix\doi{10.1109/JSTARS.2020.3021052}.
\bibitem[{Anderson et~al.(2016)Anderson, Kalra, Stanley, Sorensen, Samaras \& Oluwatola}]{RAND1}
\bibinfo{author}{Anderson, J.~M.}, \bibinfo{author}{Kalra, N.}, \bibinfo{author}{Stanley, K.~D.}, \bibinfo{author}{Sorensen, P.}, \bibinfo{author}{Samaras, C.}, \& \bibinfo{author}{Oluwatola, T.~A.} (\bibinfo{year}{2016}).
\newblock {\it \bibinfo{title}{Autonomous Vehicle Technology: A Guide for Policymakers}\/}.
\newblock \bibinfo{type}{Technical Report} RAND Corporation.
\newblock \URLprefix \url{https://www.rand.org/pubs/research_reports/RR443-2.html} \bibinfo{note}{{L}ast accessed on 2025-05-11}.
\bibitem[{Anselin(1988)}]{Anselin1988}
\bibinfo{author}{Anselin, L.} (\bibinfo{year}{1988}).
\newblock {\it \bibinfo{title}{Spatial Econometrics: Methods and Models}\/} volume~\bibinfo{volume}{4}.
\newblock \bibinfo{publisher}{Springer Netherlands}.
\newblock \DOIprefix\doi{10.1007/978-94-015-7799-1}.
\bibitem[{Bonnefon et~al.(2016)Bonnefon, Shariff \& Rahwan}]{Bonnefon2016}
\bibinfo{author}{Bonnefon, J.-F.}, \bibinfo{author}{Shariff, A.}, \& \bibinfo{author}{Rahwan, I.} (\bibinfo{year}{2016}).
\newblock \bibinfo{title}{The social dilemma of autonomous vehicles}.
\newblock {\it \bibinfo{journal}{Science}\/},  {\it \bibinfo{volume}{352}\/}, \bibinfo{pages}{1573--1576}. \DOIprefix\doi{10.1126/science.aaf2654}.
\bibitem[{Breiman(2001)}]{Breiman2001}
\bibinfo{author}{Breiman, L.} (\bibinfo{year}{2001}).
\newblock \bibinfo{title}{Random forests}.
\newblock {\it \bibinfo{journal}{Mach. Learn.}\/},  {\it \bibinfo{volume}{45}\/}, \bibinfo{pages}{5--32}. \DOIprefix\doi{10.1023/A:1010933404324}.
\bibitem[{Busch(2024)}]{Busch2024}
\bibinfo{author}{Busch, P.~A.} (\bibinfo{year}{2024}).
\newblock \bibinfo{title}{Non-user acceptance of autonomous technology: A survey of bicyclist receptivity to fully autonomous vehicles}.
\newblock {\it \bibinfo{journal}{Comput. Hum. Behav. Rep.}\/},  {\it \bibinfo{volume}{16}\/}, \bibinfo{pages}{100490}. \DOIprefix\doi{10.1016/J.CHBR.2024.100490}.
\bibitem[{{Center for Sustainable Systems}(2024)}]{UM-CSS}
\bibinfo{author}{{Center for Sustainable Systems}} (\bibinfo{year}{2024}).
\newblock \bibinfo{title}{Autonomous vehicles factsheet}.
\newblock \URLprefix \url{https://css.umich.edu/publications/factsheets/mobility/autonomous-vehicles-factsheet} \bibinfo{note}{{L}ast accessed on 2025-04-01}.
\bibitem[{Cervero(1996)}]{Cervero1996}
\bibinfo{author}{Cervero, R.} (\bibinfo{year}{1996}).
\newblock \bibinfo{title}{Mixed land-uses and commuting: Evidence from the american housing survey}.
\newblock {\it \bibinfo{journal}{Transp. Res. A}\/},  {\it \bibinfo{volume}{30}\/}, \bibinfo{pages}{361--377}. \DOIprefix\doi{10.1016/0965-8564(95)00033-X}.
\bibitem[{Cervero \& Guerra(2011)}]{Cervero2011}
\bibinfo{author}{Cervero, R.}, \& \bibinfo{author}{Guerra, E.} (\bibinfo{year}{2011}).
\newblock \bibinfo{title}{Urban densities and transit: A multi-dimensional perspective}.
\bibitem[{Cervero \& Kockelman(1997)}]{Cervero1997}
\bibinfo{author}{Cervero, R.}, \& \bibinfo{author}{Kockelman, K.} (\bibinfo{year}{1997}).
\newblock \bibinfo{title}{Travel demand and the 3ds: Density, diversity, and design}.
\newblock {\it \bibinfo{journal}{Transp. Res. D}\/},  {\it \bibinfo{volume}{2}\/}, \bibinfo{pages}{199--219}. \DOIprefix\doi{10.1016/S1361-9209(97)00009-6}.
\bibitem[{Chawla et~al.(2002)Chawla, Bowyer, Hall \& Kegelmeyer}]{Chawla2002}
\bibinfo{author}{Chawla, N.~V.}, \bibinfo{author}{Bowyer, K.~W.}, \bibinfo{author}{Hall, L.~O.}, \& \bibinfo{author}{Kegelmeyer, W.~P.} (\bibinfo{year}{2002}).
\newblock \bibinfo{title}{Smote: Synthetic minority over-sampling technique}.
\newblock {\it \bibinfo{journal}{J. Artif. Intell.}\/},  {\it \bibinfo{volume}{16}\/}, \bibinfo{pages}{321--357}. \DOIprefix\doi{10.1613/jair.953}.
\bibitem[{Chen et~al.(2020)Chen, Chen, Liu, Sun \& Zhou}]{ChenH2020}
\bibinfo{author}{Chen, H.}, \bibinfo{author}{Chen, H.}, \bibinfo{author}{Liu, Z.}, \bibinfo{author}{Sun, X.}, \& \bibinfo{author}{Zhou, R.} (\bibinfo{year}{2020}).
\newblock \bibinfo{title}{Analysis of factors affecting the severity of automated vehicle crashes using xgboost model combining poi data}.
\newblock {\it \bibinfo{journal}{J. Adv. Transp.}\/},  {\it \bibinfo{volume}{2020}\/}, \bibinfo{pages}{1--12}. \DOIprefix\doi{10.1155/2020/8881545}.
\bibitem[{Choi et~al.(2022)Choi, No, Park \& Kim}]{Choi2022}
\bibinfo{author}{Choi, J.}, \bibinfo{author}{No, W.}, \bibinfo{author}{Park, M.}, \& \bibinfo{author}{Kim, Y.} (\bibinfo{year}{2022}).
\newblock \bibinfo{title}{Inferring land use from spatialtemporal taxi ride data}.
\newblock {\it \bibinfo{journal}{Appl. Geogr.}\/},  {\it \bibinfo{volume}{142}\/}, \bibinfo{pages}{102688}. \DOIprefix\doi{10.1016/J.APGEOG.2022.102688}.
\bibitem[{Deichmann et~al.(2023)Deichmann, Ebel, Heineke, Heuss, Kellner \& Steiner}]{Deichmann2023}
\bibinfo{author}{Deichmann, J.}, \bibinfo{author}{Ebel, E.}, \bibinfo{author}{Heineke, K.}, \bibinfo{author}{Heuss, R.}, \bibinfo{author}{Kellner, M.}, \& \bibinfo{author}{Steiner, F.} (\bibinfo{year}{2023}).
\newblock \bibinfo{title}{Autonomous driving’s future: Convenient and connected}.
\newblock \URLprefix \url{https://www.mckinsey.com/industries/automotive-and-assembly/our-insights/autonomous-drivings-future-convenient-and-connected} \bibinfo{note}{{L}ast accessed on 2025-04-01}.
\bibitem[{Ding et~al.(2024)Ding, Wang, Li, Zhou, Sze \& Dong}]{Ding2024}
\bibinfo{author}{Ding, H.}, \bibinfo{author}{Wang, R.}, \bibinfo{author}{Li, T.}, \bibinfo{author}{Zhou, M.}, \bibinfo{author}{Sze, N.~N.}, \& \bibinfo{author}{Dong, N.} (\bibinfo{year}{2024}).
\newblock \bibinfo{title}{Quantifying the heterogeneity impact of risk factors on regional bicycle crash frequency: A hybrid approach of clustering and random parameter model}.
\newblock {\it \bibinfo{journal}{Accid. Anal. Prev.}\/},  {\it \bibinfo{volume}{207}\/}, \bibinfo{pages}{107753}. \DOIprefix\doi{10.1016/j.aap.2024.107753}.
\bibitem[{Du et~al.(2015)Du, Samat, Waske, Liu \& Li}]{Du2015}
\bibinfo{author}{Du, P.}, \bibinfo{author}{Samat, A.}, \bibinfo{author}{Waske, B.}, \bibinfo{author}{Liu, S.}, \& \bibinfo{author}{Li, Z.} (\bibinfo{year}{2015}).
\newblock \bibinfo{title}{Random forest and rotation forest for fully polarized sar image classification using polarimetric and spatial features}.
\newblock {\it \bibinfo{journal}{ISPRS J. Photogramm. Remote Sens.}\/},  {\it \bibinfo{volume}{105}\/}, \bibinfo{pages}{38--53}. \DOIprefix\doi{10.1016/j.isprsjprs.2015.03.002}.
\bibitem[{Dumbaugh \& Li(2010)}]{Dumbaugh2010}
\bibinfo{author}{Dumbaugh, E.}, \& \bibinfo{author}{Li, W.} (\bibinfo{year}{2010}).
\newblock \bibinfo{title}{Designing for the safety of pedestrians, cyclists, and motorists in urban environments}.
\newblock {\it \bibinfo{journal}{J. Am. Plan. Assoc.}\/},  {\it \bibinfo{volume}{77}\/}, \bibinfo{pages}{69--88}. \DOIprefix\doi{10.1080/01944363.2011.536101}.
\bibitem[{Dumbaugh \& Rae(2009)}]{Dumbaugh2009}
\bibinfo{author}{Dumbaugh, E.}, \& \bibinfo{author}{Rae, R.} (\bibinfo{year}{2009}).
\newblock \bibinfo{title}{Safe urban form: Revisiting the relationship between community design and traffic safety}.
\newblock {\it \bibinfo{journal}{J. Am. Plan. Assoc.}\/},  {\it \bibinfo{volume}{75}\/}, \bibinfo{pages}{309--329}. \DOIprefix\doi{10.1080/01944360902950349}.
\bibitem[{Ewing(2016)}]{Ewing2016}
\bibinfo{author}{Ewing, R.} (\bibinfo{year}{2016}).
\newblock \bibinfo{title}{Contribution of urban design qualities to pedestrian activity}.
\newblock {\it \bibinfo{journal}{Planning Magazine}\/}, . \URLprefix \url{https://planning.org/planning/2016/feb/research.htm}.
\newblock \bibinfo{note}{{L}ast accessed on 2025-05-11}.
\bibitem[{Ewing \& Cervero(2001)}]{Ewing2001}
\bibinfo{author}{Ewing, R.}, \& \bibinfo{author}{Cervero, R.} (\bibinfo{year}{2001}).
\newblock \bibinfo{title}{Travel and the built environment: A synthesis}.
\newblock {\it \bibinfo{journal}{Transp. Res. Rec.}\/},  {\it \bibinfo{volume}{1780}\/}, \bibinfo{pages}{87--114}. \DOIprefix\doi{10.3141/1780-10}.
\bibitem[{Ewing \& Dumbaugh(2009)}]{Ewing2009}
\bibinfo{author}{Ewing, R.}, \& \bibinfo{author}{Dumbaugh, E.} (\bibinfo{year}{2009}).
\newblock \bibinfo{title}{The built environment and traffic safety: A review of empirical evidence}.
\newblock {\it \bibinfo{journal}{J. Plan. Lit.}\/},  {\it \bibinfo{volume}{23}\/}, \bibinfo{pages}{347--367}. \DOIprefix\doi{10.1177/0885412209335553}.
\bibitem[{Fagnant \& Kockelman(2015)}]{Fagnant2015}
\bibinfo{author}{Fagnant, D.~J.}, \& \bibinfo{author}{Kockelman, K.} (\bibinfo{year}{2015}).
\newblock \bibinfo{title}{Preparing a nation for autonomous vehicles: opportunities, barriers and policy recommendations}.
\newblock {\it \bibinfo{journal}{Transp. Res. A}\/},  {\it \bibinfo{volume}{77}\/}, \bibinfo{pages}{167--181}. \DOIprefix\doi{10.1016/J.TRA.2015.04.003}.
\bibitem[{Favarò et~al.(2017)Favarò, Nader, Eurich, Tripp \& Varadaraju}]{Favar2017}
\bibinfo{author}{Favarò, F.~M.}, \bibinfo{author}{Nader, N.}, \bibinfo{author}{Eurich, S.~O.}, \bibinfo{author}{Tripp, M.}, \& \bibinfo{author}{Varadaraju, N.} (\bibinfo{year}{2017}).
\newblock \bibinfo{title}{Examining accident reports involving autonomous vehicles in california}.
\newblock {\it \bibinfo{journal}{PLOS ONE}\/},  {\it \bibinfo{volume}{12}\/}, \bibinfo{pages}{e0184952}. \DOIprefix\doi{10.1371/journal.pone.0184952}.
\bibitem[{Ferenchak \& Marshall(2024)}]{Ferenchak2024}
\bibinfo{author}{Ferenchak, N.~N.}, \& \bibinfo{author}{Marshall, W.~E.} (\bibinfo{year}{2024}).
\newblock \bibinfo{title}{Traffic safety for all road users: A paired comparison study of small \& mid-sized u.s. cities with high/low bicycling rates}.
\newblock {\it \bibinfo{journal}{J. Cycl. Micromobility Res.}\/},  {\it \bibinfo{volume}{2}\/}, \bibinfo{pages}{100010}. \DOIprefix\doi{10.1016/J.JCMR.2024.100010}.
\bibitem[{Flahaut(2004)}]{Flahaut2004}
\bibinfo{author}{Flahaut, B.} (\bibinfo{year}{2004}).
\newblock \bibinfo{title}{Impact of infrastructure and local environment on road unsafety: Logistic modeling with spatial autocorrelation}.
\newblock {\it \bibinfo{journal}{Accid. Anal. Prev.}\/},  {\it \bibinfo{volume}{36}\/}, \bibinfo{pages}{1055--1066}. \DOIprefix\doi{10.1016/J.AAP.2003.12.003}.
\bibitem[{Fu et~al.(2025)Fu, Ye, Fu, Chen \& Zhao}]{FuH2025}
\bibinfo{author}{Fu, H.}, \bibinfo{author}{Ye, S.}, \bibinfo{author}{Fu, X.}, \bibinfo{author}{Chen, T.}, \& \bibinfo{author}{Zhao, J.} (\bibinfo{year}{2025}).
\newblock \bibinfo{title}{New insights into factors affecting the severity of autonomous vehicle crashes from two sources of {AV} incident records}.
\newblock {\it \bibinfo{journal}{Travel Behav. Soc.}\/},  {\it \bibinfo{volume}{38}\/}, \bibinfo{pages}{100934}. \DOIprefix\doi{10.1016/J.TBS.2024.100934}.
\bibitem[{Gao et~al.(2023)Gao, Hu \& Li}]{Gao2023}
\bibinfo{author}{Gao, S.}, \bibinfo{author}{Hu, Y.}, \& \bibinfo{author}{Li, W.} (\bibinfo{year}{2023}).
\newblock \bibinfo{title}{Heterogeneity-aware deep learning in space: Performance and fairness}.
\newblock In {\it \bibinfo{booktitle}{Handbook of Geospatial Artificial Intelligence}\/} (pp. \bibinfo{pages}{151--176}).
\newblock \bibinfo{publisher}{CRC Press}.
\newblock \DOIprefix\doi{10.1201/9781003308423}.
\bibitem[{Georganos et~al.(2021)Georganos, Grippa, Gadiaga, Linard, Lennert, Vanhuysse, Mboga, Wolff \& Kalogirou}]{Georganos2021}
\bibinfo{author}{Georganos, S.}, \bibinfo{author}{Grippa, T.}, \bibinfo{author}{Gadiaga, A.~N.}, \bibinfo{author}{Linard, C.}, \bibinfo{author}{Lennert, M.}, \bibinfo{author}{Vanhuysse, S.}, \bibinfo{author}{Mboga, N.}, \bibinfo{author}{Wolff, E.}, \& \bibinfo{author}{Kalogirou, S.} (\bibinfo{year}{2021}).
\newblock \bibinfo{title}{Geographical random forests: a spatial extension of the random forest algorithm to address spatial heterogeneity in remote sensing and population modelling}.
\newblock {\it \bibinfo{journal}{Geocarto Int.}\/},  {\it \bibinfo{volume}{36}\/}, \bibinfo{pages}{121--136}. \DOIprefix\doi{10.1080/10106049.2019.1595177}.
\bibitem[{{Goldman Sachs}(2024)}]{GoldmanSachs2024}
\bibinfo{author}{{Goldman Sachs}} (\bibinfo{year}{2024}).
\newblock \bibinfo{title}{Partially autonomous cars forecast to comprise 10\% of new vehicle sales by 2030}.
\newblock \URLprefix \url{https://www.goldmansachs.com/insights/articles/partially-autonomous-cars-forecast-to-comprise-10-percent-of-new-vehicle-sales-by-2030} \bibinfo{note}{{L}ast accessed on 2025-04-01}.
\bibitem[{Handy(2018)}]{Handy2018}
\bibinfo{author}{Handy, S.} (\bibinfo{year}{2018}).
\newblock \bibinfo{title}{Enough with the ``ds'' already - let’s get back to ``a''}.
\newblock \URLprefix \url{https://transfersmagazine.org/magazine-article/issue-1/enough-with-the-ds-already-lets-get-back-to-a/} \bibinfo{note}{{L}ast accessed on 2025-04-07}.
\bibitem[{He \& Garcia(2009)}]{HeH2009}
\bibinfo{author}{He, H.}, \& \bibinfo{author}{Garcia, E.} (\bibinfo{year}{2009}).
\newblock \bibinfo{title}{Learning from imbalanced data}.
\newblock {\it \bibinfo{journal}{IEEE Trans. Knowl. Data Eng.}\/},  {\it \bibinfo{volume}{21}\/}, \bibinfo{pages}{1263--1284}. \DOIprefix\doi{10.1109/TKDE.2008.239}.
\bibitem[{Hepp \& Walker(2016)}]{Hepp2016}
\bibinfo{author}{Hepp, N.}, \& \bibinfo{author}{Walker, L.} (\bibinfo{year}{2016}).
\newblock \bibinfo{title}{Built environment}.
\newblock \URLprefix \url{https://www.healthandenvironment.org/resources/environmental-hazards/exposure-sources/built-environment} \bibinfo{note}{{L}ast accessed on 2025-04-04}.
\bibitem[{Huang et~al.(2018)Huang, Wang \& Patton}]{Huang2018}
\bibinfo{author}{Huang, Y.}, \bibinfo{author}{Wang, X.}, \& \bibinfo{author}{Patton, D.} (\bibinfo{year}{2018}).
\newblock \bibinfo{title}{Examining spatial relationships between crashes and the built environment: A geographically weighted regression approach}.
\newblock {\it \bibinfo{journal}{J. Transp. Geogr.}\/},  {\it \bibinfo{volume}{69}\/}, \bibinfo{pages}{221--233}. \DOIprefix\doi{10.1016/j.jtrangeo.2018.04.027}.
\bibitem[{Ignatious et~al.(2022)Ignatious, El-Sayed \& Khan}]{Ignatious2022}
\bibinfo{author}{Ignatious, H.~A.}, \bibinfo{author}{El-Sayed, H.}, \& \bibinfo{author}{Khan, M.} (\bibinfo{year}{2022}).
\newblock \bibinfo{title}{An overview of sensors in autonomous vehicles}.
\newblock {\it \bibinfo{journal}{Procedia Comput. Sci.}\/},  {\it \bibinfo{volume}{198}\/}, \bibinfo{pages}{736--741}. \DOIprefix\doi{10.1016/j.procs.2021.12.315}.
\bibitem[{James et~al.(2021)James, Witten, Hastie \& Tibshirani}]{James2021}
\bibinfo{author}{James, G.}, \bibinfo{author}{Witten, D.}, \bibinfo{author}{Hastie, T.}, \& \bibinfo{author}{Tibshirani, R.} (\bibinfo{year}{2021}).
\newblock \bibinfo{title}{Linear regression}.
\newblock In {\it \bibinfo{booktitle}{An Introduction to Statistical Learning}\/} (pp. \bibinfo{pages}{59--128}).
\newblock \bibinfo{publisher}{Springer}. (\bibinfo{edition}{2nd} ed.).
\newblock \DOIprefix\doi{10.1007/978-1-0716-1418-1_3}.
\bibitem[{Johnson \& Iizuka(2016)}]{Johnson2016}
\bibinfo{author}{Johnson, B.~A.}, \& \bibinfo{author}{Iizuka, K.} (\bibinfo{year}{2016}).
\newblock \bibinfo{title}{Integrating openstreetmap crowdsourced data and landsat time-series imagery for rapid land use/land cover (lulc) mapping: Case study of the laguna de bay area of the philippines}.
\newblock {\it \bibinfo{journal}{Appl. Geogr.}\/},  {\it \bibinfo{volume}{67}\/}, \bibinfo{pages}{140--149}. \DOIprefix\doi{10.1016/j.apgeog.2015.12.006}.
\bibitem[{Kim \& Park(2019)}]{Kim2019}
\bibinfo{author}{Kim, T.~K.}, \& \bibinfo{author}{Park, J.~H.} (\bibinfo{year}{2019}).
\newblock \bibinfo{title}{More about the basic assumptions of t-test: normality and sample size}.
\newblock {\it \bibinfo{journal}{Korean J. Anesthesiol.}\/},  {\it \bibinfo{volume}{72}\/}, \bibinfo{pages}{331--335}. \DOIprefix\doi{10.4097/kja.d.18.00292}.
\bibitem[{Krawczyk(2016)}]{Krawczyk2016}
\bibinfo{author}{Krawczyk, B.} (\bibinfo{year}{2016}).
\newblock \bibinfo{title}{Learning from imbalanced data: open challenges and future directions}.
\newblock {\it \bibinfo{journal}{Prog. Artif. Intell.}\/},  {\it \bibinfo{volume}{5}\/}, \bibinfo{pages}{221--232}. \DOIprefix\doi{10.1007/s13748-016-0094-0}.
\bibitem[{Kuo et~al.(2024{\natexlab{a}})Kuo, Hsu, Lord \& Putra}]{Kuo2024b}
\bibinfo{author}{Kuo, P.-F.}, \bibinfo{author}{Hsu, W.-T.}, \bibinfo{author}{Lord, D.}, \& \bibinfo{author}{Putra, I. G.~B.} (\bibinfo{year}{2024}{\natexlab{a}}).
\newblock \bibinfo{title}{Classification of autonomous vehicle crash severity: Solving the problems of imbalanced datasets and small sample size}.
\newblock {\it \bibinfo{journal}{Accid. Anal. Prev.}\/},  {\it \bibinfo{volume}{205}\/}, \bibinfo{pages}{107666}. \DOIprefix\doi{10.1016/j.aap.2024.107666}.
\bibitem[{Kuo et~al.(2024{\natexlab{b}})Kuo, Sulistyah, Putra \& Lord}]{Kuo2024a}
\bibinfo{author}{Kuo, P.-F.}, \bibinfo{author}{Sulistyah, U.~D.}, \bibinfo{author}{Putra, I. G.~B.}, \& \bibinfo{author}{Lord, D.} (\bibinfo{year}{2024}{\natexlab{b}}).
\newblock \bibinfo{title}{Exploring the spatial relationship of e-bike and motorcycle crashes: Implications for risk reduction}.
\newblock {\it \bibinfo{journal}{J. Safety. Res.}\/},  {\it \bibinfo{volume}{88}\/}, \bibinfo{pages}{199--216}. \DOIprefix\doi{10.1016/j.jsr.2023.11.007}.
\bibitem[{Kurse et~al.(2025)Kurse, Gebresenbet, Daba \& Tefera}]{Kurse2025}
\bibinfo{author}{Kurse, T.~K.}, \bibinfo{author}{Gebresenbet, G.}, \bibinfo{author}{Daba, G.~F.}, \& \bibinfo{author}{Tefera, N.~T.} (\bibinfo{year}{2025}).
\newblock \bibinfo{title}{Experimental determination of factors causing crashes involving automated vehicles}.
\newblock {\it \bibinfo{journal}{Multimodal Transp.}\/},  {\it \bibinfo{volume}{4}\/}, \bibinfo{pages}{100186}. \DOIprefix\doi{10.1016/J.MULTRA.2024.100186}.
\bibitem[{Kutela et~al.(2022)Kutela, Das \& Dadashova}]{Kutela2022}
\bibinfo{author}{Kutela, B.}, \bibinfo{author}{Das, S.}, \& \bibinfo{author}{Dadashova, B.} (\bibinfo{year}{2022}).
\newblock \bibinfo{title}{Mining patterns of autonomous vehicle crashes involving vulnerable road users to understand the associated factors}.
\newblock {\it \bibinfo{journal}{Accid. Anal. Prev.}\/},  {\it \bibinfo{volume}{165}\/}, \bibinfo{pages}{106473}. \DOIprefix\doi{10.1016/J.AAP.2021.106473}.
\bibitem[{LeSage \& Pace(2009)}]{LeSage2009}
\bibinfo{author}{LeSage, J.}, \& \bibinfo{author}{Pace, R.~K.} (\bibinfo{year}{2009}).
\newblock \bibinfo{title}{Introduction}.
\newblock In {\it \bibinfo{booktitle}{Introduction to Spatial Econometrics}\/} (pp. \bibinfo{pages}{1--24}).
\newblock \bibinfo{publisher}{Chapman and Hall/CRC}.
\newblock \DOIprefix\doi{10.1201/9781420064254}.
\bibitem[{Li et~al.(2024)Li, Chen, Yue, Xu \& Noyce}]{LiP2024}
\bibinfo{author}{Li, P.}, \bibinfo{author}{Chen, S.}, \bibinfo{author}{Yue, L.}, \bibinfo{author}{Xu, Y.}, \& \bibinfo{author}{Noyce, D.~A.} (\bibinfo{year}{2024}).
\newblock \bibinfo{title}{Analyzing relationships between latent topics in autonomous vehicle crash narratives and crash severity using natural language processing techniques and explainable xgboost}.
\newblock {\it \bibinfo{journal}{Accid. Anal. Prev.}\/},  {\it \bibinfo{volume}{203}\/}, \bibinfo{pages}{107605}. \DOIprefix\doi{10.1016/j.aap.2024.107605}.
\bibitem[{Li et~al.(2025{\natexlab{a}})Li, Liu, Fan, Zhao, Zhang, Fan \& Li}]{LiT2025}
\bibinfo{author}{Li, T.}, \bibinfo{author}{Liu, S.}, \bibinfo{author}{Fan, G.}, \bibinfo{author}{Zhao, H.}, \bibinfo{author}{Zhang, M.}, \bibinfo{author}{Fan, J.}, \& \bibinfo{author}{Li, C.} (\bibinfo{year}{2025}{\natexlab{a}}).
\newblock \bibinfo{title}{Spatial heterogeneity effect of built environment on traffic safety using geographically weighted atrous convolutions neural network}.
\newblock {\it \bibinfo{journal}{Accid. Anal. Prev.}\/},  {\it \bibinfo{volume}{213}\/}, \bibinfo{pages}{107934}. \DOIprefix\doi{10.1016/j.aap.2025.107934}.
\bibitem[{Li et~al.(2025{\natexlab{b}})Li, Wang, Wang \& Liu}]{LiY2025}
\bibinfo{author}{Li, Y.}, \bibinfo{author}{Wang, X.}, \bibinfo{author}{Wang, T.}, \& \bibinfo{author}{Liu, Q.} (\bibinfo{year}{2025}{\natexlab{b}}).
\newblock \bibinfo{title}{Characteristics analysis of autonomous vehicle pre-crash scenarios}.
\newblock \URLprefix \url{https://arxiv.org/abs/2502.20789}. \href{http://arxiv.org/abs/2502.20789}{\tt arXiv:2502.20789} \bibinfo{note}{arXiv preprint}.
\bibitem[{Liu et~al.(2024)Liu, Zhang, Cheng \& Wang}]{LiuH2024}
\bibinfo{author}{Liu, H.}, \bibinfo{author}{Zhang, W.}, \bibinfo{author}{Cheng, Z.}, \& \bibinfo{author}{Wang, T.} (\bibinfo{year}{2024}).
\newblock \bibinfo{title}{Investigating the contributing factors of urban crash levels: A novel stacking integrated learning framework}.
\newblock {\it \bibinfo{journal}{Appl. Geogr.}\/},  {\it \bibinfo{volume}{173}\/}, \bibinfo{pages}{103440}. \DOIprefix\doi{10.1016/J.APGEOG.2024.103440}.
\bibitem[{Liu et~al.(2023)Liu, Chen, Tian \& Vos}]{LiuX2023}
\bibinfo{author}{Liu, X.}, \bibinfo{author}{Chen, X.}, \bibinfo{author}{Tian, M.}, \& \bibinfo{author}{Vos, J.~D.} (\bibinfo{year}{2023}).
\newblock \bibinfo{title}{Effects of buffer size on associations between the built environment and metro ridership: A machine learning-based sensitive analysis}.
\newblock {\it \bibinfo{journal}{J. Transp. Geogr.}\/},  {\it \bibinfo{volume}{113}\/}, \bibinfo{pages}{103730}. \DOIprefix\doi{10.1016/j.jtrangeo.2023.103730}.
\bibitem[{Liu et~al.(2025)Liu, Chen, Chung, Jang \& Xu}]{LiuY2025}
\bibinfo{author}{Liu, Y.}, \bibinfo{author}{Chen, T.}, \bibinfo{author}{Chung, H.}, \bibinfo{author}{Jang, K.}, \& \bibinfo{author}{Xu, P.} (\bibinfo{year}{2025}).
\newblock \bibinfo{title}{Is there an emotional dimension to road safety? a spatial analysis for traffic crashes considering streetscape perception and built environment}.
\newblock {\it \bibinfo{journal}{Anal. Methods Accid. Res.}\/},  {\it \bibinfo{volume}{46}\/}, \bibinfo{pages}{100374}. \DOIprefix\doi{10.1016/j.amar.2025.100374}.
\bibitem[{Lu et~al.(2025)Lu, Yang, Yuan \& Peng}]{LuW2025}
\bibinfo{author}{Lu, W.}, \bibinfo{author}{Yang, B.}, \bibinfo{author}{Yuan, L.}, \& \bibinfo{author}{Peng, Z.} (\bibinfo{year}{2025}).
\newblock \bibinfo{title}{Understanding fly-tipping in urban areas: A social-economic-spatial combinatorial approach enabled by geographically weighted random forest}.
\newblock {\it \bibinfo{journal}{Environ. Impact Assess. Rev.}\/},  {\it \bibinfo{volume}{112}\/}, \bibinfo{pages}{107858}. \DOIprefix\doi{10.1016/j.eiar.2025.107858}.
\bibitem[{Mai et~al.(2025)Mai, Xie, Jia, Lao, Rao, Zhu, Liu, Chiang \& Jiao}]{Mai2025}
\bibinfo{author}{Mai, G.}, \bibinfo{author}{Xie, Y.}, \bibinfo{author}{Jia, X.}, \bibinfo{author}{Lao, N.}, \bibinfo{author}{Rao, J.}, \bibinfo{author}{Zhu, Q.}, \bibinfo{author}{Liu, Z.}, \bibinfo{author}{Chiang, Y.-Y.}, \& \bibinfo{author}{Jiao, J.} (\bibinfo{year}{2025}).
\newblock \bibinfo{title}{Towards the next generation of geospatial artificial intelligence}.
\newblock {\it \bibinfo{journal}{Int. J. Appl. Earth Obs. Geoinf.}\/},  {\it \bibinfo{volume}{136}\/}, \bibinfo{pages}{104368}. \DOIprefix\doi{10.1016/j.jag.2025.104368}.
\bibitem[{Marshall \& Garrick(2011)}]{Marshall2011}
\bibinfo{author}{Marshall, W.~E.}, \& \bibinfo{author}{Garrick, N.~W.} (\bibinfo{year}{2011}).
\newblock \bibinfo{title}{Does street network design affect traffic safety?}
\newblock {\it \bibinfo{journal}{Accid. Anal. Prev.}\/},  {\it \bibinfo{volume}{43}\/}, \bibinfo{pages}{769--781}. \DOIprefix\doi{10.1016/j.aap.2010.10.024}.
\bibitem[{Mehaffy et~al.(2015)Mehaffy, Porta \& Romice}]{Mehaffy2015}
\bibinfo{author}{Mehaffy, M.~W.}, \bibinfo{author}{Porta, S.}, \& \bibinfo{author}{Romice, O.} (\bibinfo{year}{2015}).
\newblock \bibinfo{title}{The “neighborhood unit” on trial: a case study in the impacts of urban morphology}.
\newblock {\it \bibinfo{journal}{J. Urban. Int. Res. Placemaking Urban Sustain.}\/},  {\it \bibinfo{volume}{8}\/}, \bibinfo{pages}{199--217}. \DOIprefix\doi{10.1080/17549175.2014.908786}.
\bibitem[{Merlin et~al.(2020)Merlin, Guerra \& Dumbaugh}]{Merlin2020}
\bibinfo{author}{Merlin, L.~A.}, \bibinfo{author}{Guerra, E.}, \& \bibinfo{author}{Dumbaugh, E.} (\bibinfo{year}{2020}).
\newblock \bibinfo{title}{Crash risk, crash exposure, and the built environment: A conceptual review}.
\newblock {\it \bibinfo{journal}{Accid. Anal. Prev.}\/},  {\it \bibinfo{volume}{134}\/}, \bibinfo{pages}{105244}. \DOIprefix\doi{10.1016/J.AAP.2019.07.020}.
\bibitem[{Montgomery \& Runger(2018)}]{Montgomery2018}
\bibinfo{author}{Montgomery, D.~C.}, \& \bibinfo{author}{Runger, G.~C.} (\bibinfo{year}{2018}).
\newblock \bibinfo{title}{Nonparametric procedures}.
\newblock In {\it \bibinfo{booktitle}{Applied Statistics and Probability for Engineers}\/} (pp. \bibinfo{pages}{234--240}).
\newblock \bibinfo{publisher}{Wiley}. (\bibinfo{edition}{7th} ed.).
\bibitem[{NHTSA(2025)}]{NHTSA-DATA}
\bibinfo{author}{NHTSA} (\bibinfo{year}{2025}).
\newblock \bibinfo{title}{{ADS} incident report data}.
\newblock \URLprefix \url{https://www.nhtsa.gov/laws-regulations/standing-general-order-crash-reporting#data} \bibinfo{note}{{L}ast accessed on 2025-03-31}.
\bibitem[{NHTSA(n.d.)}]{NHTSA}
\bibinfo{author}{NHTSA} (\bibinfo{year}{n.d.}).
\newblock \bibinfo{title}{Automated vehicles for safety}.
\newblock \URLprefix \url{https://www.nhtsa.gov/vehicle-safety/automated-vehicles-safety} \bibinfo{note}{{L}ast accessed on 2025-04-01}.
\bibitem[{Osama \& Sayed(2017)}]{Osama2017}
\bibinfo{author}{Osama, A.}, \& \bibinfo{author}{Sayed, T.} (\bibinfo{year}{2017}).
\newblock \bibinfo{title}{Evaluating the impact of socioeconomics, land use, built environment, and road facility on cyclist safety}.
\newblock {\it \bibinfo{journal}{Transp. Res. Rec.}\/},  {\it \bibinfo{volume}{2659}\/}, \bibinfo{pages}{33--42}. \DOIprefix\doi{10.3141/2659-04}.
\bibitem[{Pan et~al.(2024)Pan, Zhang, Liu, Head, Elli \& Alvarez}]{Pan2024}
\bibinfo{author}{Pan, F.}, \bibinfo{author}{Zhang, Y.}, \bibinfo{author}{Liu, J.}, \bibinfo{author}{Head, L.}, \bibinfo{author}{Elli, M.}, \& \bibinfo{author}{Alvarez, I.} (\bibinfo{year}{2024}).
\newblock \bibinfo{title}{Reliability modeling for perception systems in autonomous vehicles: A recursive event-triggering point process approach}.
\newblock {\it \bibinfo{journal}{Transp. Res. C}\/},  {\it \bibinfo{volume}{169}\/}, \bibinfo{pages}{104868}. \DOIprefix\doi{10.1016/j.trc.2024.104868}.
\bibitem[{Park \& Villalobos(2024)}]{RAND2}
\bibinfo{author}{Park, H.~M.}, \& \bibinfo{author}{Villalobos, F.} (\bibinfo{year}{2024}).
\newblock \bibinfo{title}{Reining in the risks of robotaxis}.
\newblock \URLprefix \url{https://www.governing.com/policy/reining-in-the-risks-of-robotaxis} \bibinfo{note}{{L}ast accessed on 2025-03-31}.
\bibitem[{Perry(1929)}]{Perry1929}
\bibinfo{author}{Perry, C.~A.} (\bibinfo{year}{1929}).
\newblock \bibinfo{title}{The neighborhood unit: A scheme of arrangement for the family life community}.
\newblock In {\it \bibinfo{booktitle}{Regional Plan for New York and Its Environs}\/}.
\newblock volume~\bibinfo{volume}{7}.
\bibitem[{Portella(2014)}]{Portella2014}
\bibinfo{author}{Portella, A.~A.} (\bibinfo{year}{2014}).
\newblock \bibinfo{title}{Built environment}.
\newblock In {\it \bibinfo{booktitle}{Encyclopedia of Quality of Life and Well-Being Research}\/} (pp. \bibinfo{pages}{454--461}).
\newblock \bibinfo{publisher}{Springer Netherlands}.
\newblock \DOIprefix\doi{10.1007/978-94-007-0753-5_240}.
\bibitem[{PSRC(2015)}]{PSRC2015}
\bibinfo{author}{PSRC} (\bibinfo{year}{2015}).
\newblock \bibinfo{title}{Transit-supportive densities and land uses}.
\newblock \URLprefix \url{https://www.psrc.org/sites/default/files/2022-03/tsdluguidancepaper.pdf} \bibinfo{note}{{L}ast accessed on 2025-04-14}.
\bibitem[{Ren et~al.(2022)Ren, Yu, Chen \& Gao}]{Ren2022}
\bibinfo{author}{Ren, W.}, \bibinfo{author}{Yu, B.}, \bibinfo{author}{Chen, Y.}, \& \bibinfo{author}{Gao, K.} (\bibinfo{year}{2022}).
\newblock \bibinfo{title}{Divergent effects of factors on crash severity under autonomous and conventional driving modes using a hierarchical bayesian approach}.
\newblock {\it \bibinfo{journal}{Int. J. Environ. Res. Public Health}\/},  {\it \bibinfo{volume}{19}\/}, \bibinfo{pages}{11358}. \DOIprefix\doi{10.3390/ijerph191811358}.
\bibitem[{{SAE International}(2021)}]{SAE-STD}
\bibinfo{author}{{SAE International}} (\bibinfo{year}{2021}).
\newblock \bibinfo{title}{Taxonomy and definitions for terms related to driving automation systems for on-road motor vehicles}.
\newblock \URLprefix \url{https://www.sae.org/blog/sae-j3016-update} \bibinfo{note}{{L}ast accessed on 2025-05-11}.
\bibitem[{Sailaja et~al.(2024)Sailaja, Gayatri, Rathod, Padmavathi, Kumar, Kumar \& Sundaram}]{Sailaja2024}
\bibinfo{author}{Sailaja, B.}, \bibinfo{author}{Gayatri, S.}, \bibinfo{author}{Rathod, S.}, \bibinfo{author}{Padmavathi, C.}, \bibinfo{author}{Kumar, R.~N.}, \bibinfo{author}{Kumar, R.~M.}, \& \bibinfo{author}{Sundaram, R.~M.} (\bibinfo{year}{2024}).
\newblock \bibinfo{title}{Spatial temperature prediction—a machine learning and gis perspective}.
\newblock {\it \bibinfo{journal}{Theor. Appl. Climatol.}\/},  {\it \bibinfo{volume}{155}\/}, \bibinfo{pages}{9619--9642}. \DOIprefix\doi{10.1007/s00704-024-05167-3}.
\bibitem[{Sarkar et~al.(2025)Sarkar, Biswas, Koramati, Sinha \& Majumdar}]{Sarkar2025}
\bibinfo{author}{Sarkar, S.}, \bibinfo{author}{Biswas, S.}, \bibinfo{author}{Koramati, S.}, \bibinfo{author}{Sinha, A.~K.}, \& \bibinfo{author}{Majumdar, B.~B.} (\bibinfo{year}{2025}).
\newblock \bibinfo{title}{Determination of various risk elements’ association causing fatal crashes in heterogeneous traffic conditions}.
\newblock {\it \bibinfo{journal}{Transp. Res. Rec.}\/}, . \DOIprefix\doi{10.1177/03611981241312221}.
\bibitem[{Sun et~al.(2024)Sun, Zhou, Kim \& Hu}]{Sun2024}
\bibinfo{author}{Sun, K.}, \bibinfo{author}{Zhou, R.~Z.}, \bibinfo{author}{Kim, J.}, \& \bibinfo{author}{Hu, Y.} (\bibinfo{year}{2024}).
\newblock \bibinfo{title}{{PyGRF}: An improved python geographical random forest model and case studies in public health and natural disasters}.
\newblock {\it \bibinfo{journal}{Trans. GIS}\/},  {\it \bibinfo{volume}{28}\/}, \bibinfo{pages}{2476--2491}. \DOIprefix\doi{10.1111/tgis.13248}.
\bibitem[{Ukkusuri et~al.(2012)Ukkusuri, Miranda-Moreno, Ramadurai \& Isa-Tavarez}]{Ukkusuri2012}
\bibinfo{author}{Ukkusuri, S.}, \bibinfo{author}{Miranda-Moreno, L.~F.}, \bibinfo{author}{Ramadurai, G.}, \& \bibinfo{author}{Isa-Tavarez, J.} (\bibinfo{year}{2012}).
\newblock \bibinfo{title}{The role of built environment on pedestrian crash frequency}.
\newblock {\it \bibinfo{journal}{Saf. Sci.}\/},  {\it \bibinfo{volume}{50}\/}, \bibinfo{pages}{1141--1151}. \DOIprefix\doi{10.1016/J.SSCI.2011.09.012}.
\bibitem[{{United States Environmental Protection Agency}(2025{\natexlab{a}})}]{EPA-BE}
\bibinfo{author}{{United States Environmental Protection Agency}} (\bibinfo{year}{2025}{\natexlab{a}}).
\newblock \bibinfo{title}{Basic information about the built environment}.
\newblock \URLprefix \url{https://www.epa.gov/smm/basic-information-about-built-environment} \bibinfo{note}{{L}ast accessed on 2025-04-04}.
\bibitem[{{United States Environmental Protection Agency}(2025{\natexlab{b}})}]{EPA-LU}
\bibinfo{author}{{United States Environmental Protection Agency}} (\bibinfo{year}{2025}{\natexlab{b}}).
\newblock \bibinfo{title}{Land use}.
\newblock \URLprefix \url{https://www.epa.gov/report-environment/land-use} \bibinfo{note}{{L}ast accessed on 2025-03-30}.
\bibitem[{Vargas et~al.(2021)Vargas, Alsweiss, Toker, Razdan \& Santos}]{Vargas2021}
\bibinfo{author}{Vargas, J.}, \bibinfo{author}{Alsweiss, S.}, \bibinfo{author}{Toker, O.}, \bibinfo{author}{Razdan, R.}, \& \bibinfo{author}{Santos, J.} (\bibinfo{year}{2021}).
\newblock \bibinfo{title}{An overview of autonomous vehicles sensors and their vulnerability to weather conditions}.
\newblock {\it \bibinfo{journal}{Sensors}\/},  {\it \bibinfo{volume}{21}\/}, \bibinfo{pages}{5397}. \DOIprefix\doi{10.3390/s21165397}.
\bibitem[{Wang et~al.(2024)Wang, Gao, Zhang, Yu \& Easa}]{Wang2024}
\bibinfo{author}{Wang, S.}, \bibinfo{author}{Gao, K.}, \bibinfo{author}{Zhang, L.}, \bibinfo{author}{Yu, B.}, \& \bibinfo{author}{Easa, S.~M.} (\bibinfo{year}{2024}).
\newblock \bibinfo{title}{Geographically weighted machine learning for modeling spatial heterogeneity in traffic crash frequency and determinants in us}.
\newblock {\it \bibinfo{journal}{Accid. Anal. Prev.}\/},  {\it \bibinfo{volume}{199}\/}, \bibinfo{pages}{107528}. \DOIprefix\doi{10.1016/j.aap.2024.107528}.
\bibitem[{Wong(2009)}]{Wong2009}
\bibinfo{author}{Wong, D.} (\bibinfo{year}{2009}).
\newblock \bibinfo{title}{Modifiable areal unit problem}.
\newblock In {\it \bibinfo{booktitle}{International Encyclopedia of Human Geography}\/} (pp. \bibinfo{pages}{169--174}).
\newblock \bibinfo{publisher}{Elsevier}.
\newblock \DOIprefix\doi{10.1016/B978-008044910-4.00475-2}.
\bibitem[{Wu et~al.(2024)Wu, Zhang \& Xiang}]{Wu2024}
\bibinfo{author}{Wu, D.}, \bibinfo{author}{Zhang, Y.}, \& \bibinfo{author}{Xiang, Q.} (\bibinfo{year}{2024}).
\newblock \bibinfo{title}{Geographically weighted random forests for macro-level crash frequency prediction}.
\newblock {\it \bibinfo{journal}{Accid. Anal. Prev.}\/},  {\it \bibinfo{volume}{194}\/}, \bibinfo{pages}{107370}. \DOIprefix\doi{10.1016/j.aap.2023.107370}.
\bibitem[{Wu et~al.(2021)Wu, Song \& Meng}]{Wu2021}
\bibinfo{author}{Wu, P.}, \bibinfo{author}{Song, L.}, \& \bibinfo{author}{Meng, X.} (\bibinfo{year}{2021}).
\newblock \bibinfo{title}{Influence of built environment and roadway characteristics on the frequency of vehicle crashes caused by driver inattention: A comparison between rural roads and urban roads}.
\newblock {\it \bibinfo{journal}{J. Safety. Res.}\/},  {\it \bibinfo{volume}{79}\/}, \bibinfo{pages}{199--210}. \DOIprefix\doi{10.1016/J.JSR.2021.09.001}.
\bibitem[{Xiao et~al.(2024)Xiao, Ding, Sze \& Zheng}]{XiaoD2024}
\bibinfo{author}{Xiao, D.}, \bibinfo{author}{Ding, H.}, \bibinfo{author}{Sze, N.~N.}, \& \bibinfo{author}{Zheng, N.} (\bibinfo{year}{2024}).
\newblock \bibinfo{title}{Investigating built environment and traffic flow impact on crash frequency in urban road networks}.
\newblock {\it \bibinfo{journal}{Accid. Anal. Prev.}\/},  {\it \bibinfo{volume}{201}\/}, \bibinfo{pages}{107561}. \DOIprefix\doi{10.1016/j.aap.2024.107561}.
\bibitem[{Xu et~al.(2019)Xu, Ding, Wang \& Li}]{Xu2019}
\bibinfo{author}{Xu, C.}, \bibinfo{author}{Ding, Z.}, \bibinfo{author}{Wang, C.}, \& \bibinfo{author}{Li, Z.} (\bibinfo{year}{2019}).
\newblock \bibinfo{title}{Statistical analysis of the patterns and characteristics of connected and autonomous vehicle involved crashes}.
\newblock {\it \bibinfo{journal}{J. Safety. Res.}\/},  {\it \bibinfo{volume}{71}\/}, \bibinfo{pages}{41--47}. \DOIprefix\doi{10.1016/j.jsr.2019.09.001}.
\bibitem[{Yang et~al.(2018)Yang, Franz, Zhu, Mahmoudi, Nasri \& Zhang}]{Yang2018}
\bibinfo{author}{Yang, Z.}, \bibinfo{author}{Franz, M.~L.}, \bibinfo{author}{Zhu, S.}, \bibinfo{author}{Mahmoudi, J.}, \bibinfo{author}{Nasri, A.}, \& \bibinfo{author}{Zhang, L.} (\bibinfo{year}{2018}).
\newblock \bibinfo{title}{Analysis of washington, dc taxi demand using gps and land-use data}.
\newblock {\it \bibinfo{journal}{J. Transp. Geogr.}\/},  {\it \bibinfo{volume}{66}\/}, \bibinfo{pages}{35--44}. \DOIprefix\doi{10.1016/J.JTRANGEO.2017.10.021}.
\bibitem[{Zhang et~al.(2019)Zhang, Ge, Tian \& Liou}]{Zhang2019}
\bibinfo{author}{Zhang, Y.}, \bibinfo{author}{Ge, T.}, \bibinfo{author}{Tian, W.}, \& \bibinfo{author}{Liou, Y.-A.} (\bibinfo{year}{2019}).
\newblock \bibinfo{title}{Debris flow susceptibility mapping using machine-learning techniques in shigatse area, china}.
\newblock {\it \bibinfo{journal}{Remote Sens.}\/},  {\it \bibinfo{volume}{11}\/}, \bibinfo{pages}{2801}. \DOIprefix\doi{10.3390/rs11232801}.
\bibitem[{Zhang et~al.(2025)Zhang, Li, Chen, Sze, Yang, Zhang \& Ren}]{Zhang2025}
\bibinfo{author}{Zhang, Z.}, \bibinfo{author}{Li, H.}, \bibinfo{author}{Chen, T.}, \bibinfo{author}{Sze, N.}, \bibinfo{author}{Yang, W.}, \bibinfo{author}{Zhang, Y.}, \& \bibinfo{author}{Ren, G.} (\bibinfo{year}{2025}).
\newblock \bibinfo{title}{Decision-making of autonomous vehicles in interactions with jaywalkers: A risk-aware deep reinforcement learning approach}.
\newblock {\it \bibinfo{journal}{Accid. Anal. Prev.}\/},  {\it \bibinfo{volume}{210}\/}, \bibinfo{pages}{107843}. \DOIprefix\doi{10.1016/j.aap.2024.107843}.
\bibitem[{Zhu \& Meng(2022)}]{Zhu2022}
\bibinfo{author}{Zhu, S.}, \& \bibinfo{author}{Meng, Q.} (\bibinfo{year}{2022}).
\newblock \bibinfo{title}{What can we learn from autonomous vehicle collision data on crash severity? a cost-sensitive cart approach}.
\newblock {\it \bibinfo{journal}{Accid. Anal. Prev.}\/},  {\it \bibinfo{volume}{174}\/}, \bibinfo{pages}{106769}. \DOIprefix\doi{10.1016/j.aap.2022.106769}.

\end{thebibliography}
}
\end{FlushLeft}

\end{document}